\documentclass[runningheads]{llncs}

\usepackage{eccv}

\usepackage{eccvabbrv}

\usepackage{graphicx}
\usepackage{booktabs}

\usepackage[accsupp]{axessibility}  

\usepackage[pagebackref,breaklinks,colorlinks,allcolors=eccvblue]{hyperref}
\usepackage[numbers,square,comma,sort&compress]{natbib}

\usepackage{orcidlink}

\usepackage{wrapfig}

\usepackage[table, dvipsnames]{xcolor}

\definecolor{pyblue}{HTML}{1F77B4}
\usepackage{pgfplots}
\usepackage{pgfplotstable}
\pgfplotsset{compat=1.18}
\usepackage{multirow}
\usepackage{pifont}
\usepackage{nicematrix}  

\usepackage{tikz} 
\usepackage{svg}

\usepackage[normalem]{ulem}
\useunder{\uline}{\ul}{}

\usepackage{xspace}
\makeatletter
\DeclareRobustCommand\onedot{\futurelet\@let@token\@onedot}
\def\@onedot{\ifx\@let@token.\else.\null\fi\xspace}

\def\eg{\emph{e.g}\onedot} 
\def\ie{\emph{i.e}\onedot} 
 
 \def\vs{\emph{vs}\onedot}
 
\def\etal{\emph{et al}\onedot}
\makeatother

\newcommand{\ours}{RAF\xspace}

\begin{document}

\title{Retrieval-Augmented Gaussian Avatars: Improving Expression Generalization} 

\titlerunning{Retrieval-Augmented Gaussian Avatars}

\author{Matan Levy$^1$, Gavriel Habib$^2$, Issar Tzachor$^2$, Dvir Samuel$^{3,5}$, Rami Ben-Ari$^2$,\\ Nir Darshan$^2$, Or Litany$^{4,5}$, Dani Lischinski$^1$}

\authorrunning{Levy et al.}

\institute{$^1$The Hebrew University of Jerusalem, Israel \quad $^2$OriginAI, Israel \\ $^3$Bar-Ilan University, Israel \quad $^4$Technion, Israel \quad$^5$NVIDIA
}

\maketitle

\begin{abstract}
Template-free animatable head avatars can achieve high visual fidelity by learning expression-dependent facial deformation directly from a subject’s capture, avoiding parametric face templates and hand-designed blendshape spaces. However, since learned deformation is supervised only by the expressions observed for a single identity, these models suffer from limited expression coverage and often struggle when driven by motions that deviate from the training distribution. We introduce RAF (Retrieval-Augmented Faces), a simple training-time augmentation designed for template-free head avatars that learn deformation from data. RAF constructs a large unlabeled expression bank and, during training, replaces a subset of the subject’s expression features with nearest-neighbor expressions retrieved from this bank while still reconstructing the subject’s original frames. This exposes the deformation field to a broader range of expression conditions, encouraging stronger identity–expression decoupling and improving robustness to expression distribution shift without requiring paired cross-identity data, additional annotations, or architectural changes.  
We further analyze how retrieval augmentation increases expression diversity and validate retrieval quality with a user study showing that retrieved neighbors are perceptually closer in expression and pose. Experiments on the NeRSemble benchmark demonstrate that RAF consistently improves expression fidelity over the baseline, in both self-driving and cross-driving scenarios.
\end{abstract}

\begin{figure*}[ht]
    \centering
    \resizebox{0.80\textwidth}{!}{
    \begin{minipage}{\textwidth}
    
    \begin{subfigure}[b]{0.24\textwidth}
         \centering
         \includegraphics[width=\textwidth]{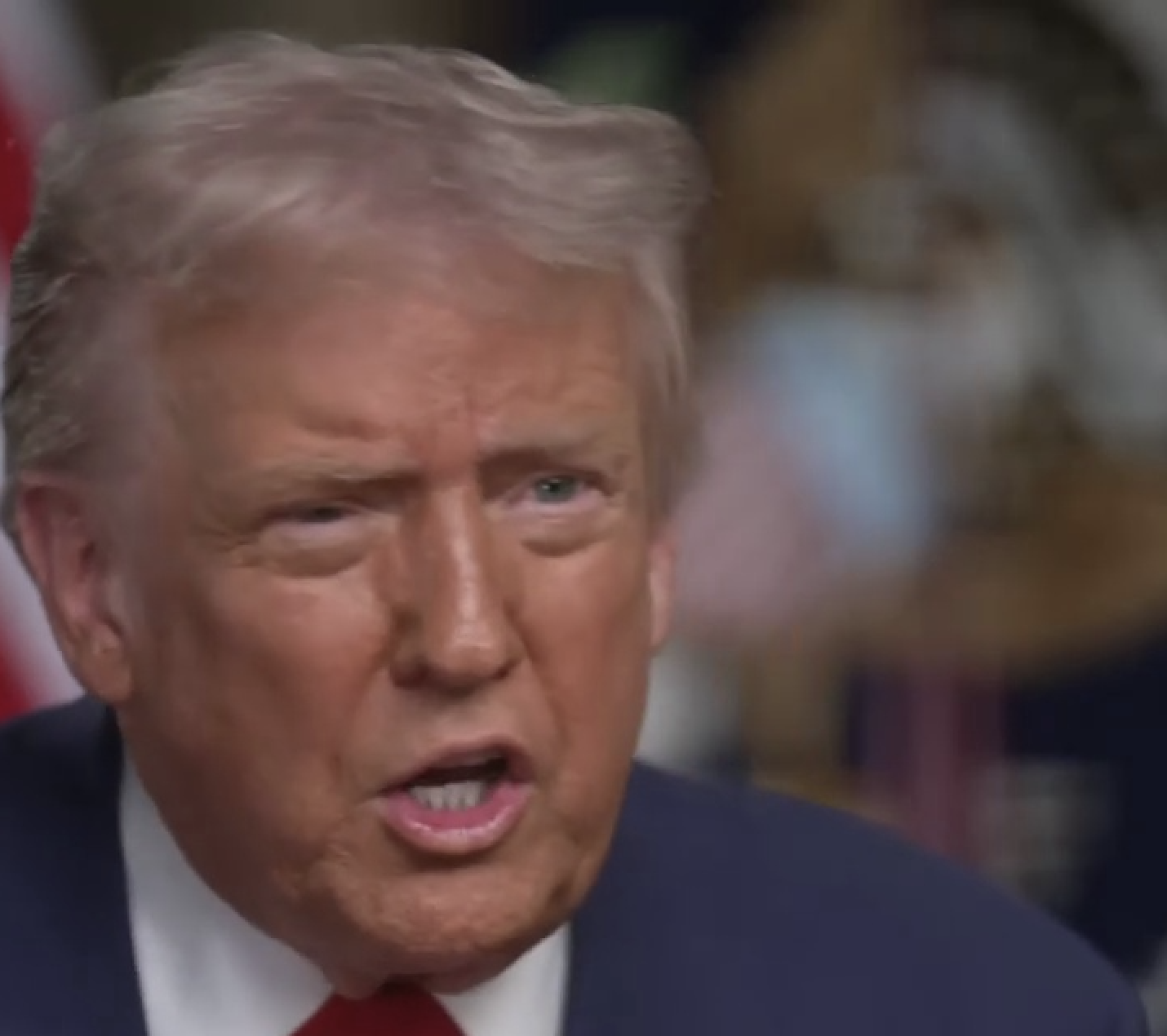}
         \caption{Driver}
         \label{fig:example_d445_a443_a}
     \end{subfigure}
     \hfil
     \begin{subfigure}[b]{0.24\textwidth}
         \centering
         \includegraphics[width=\textwidth]{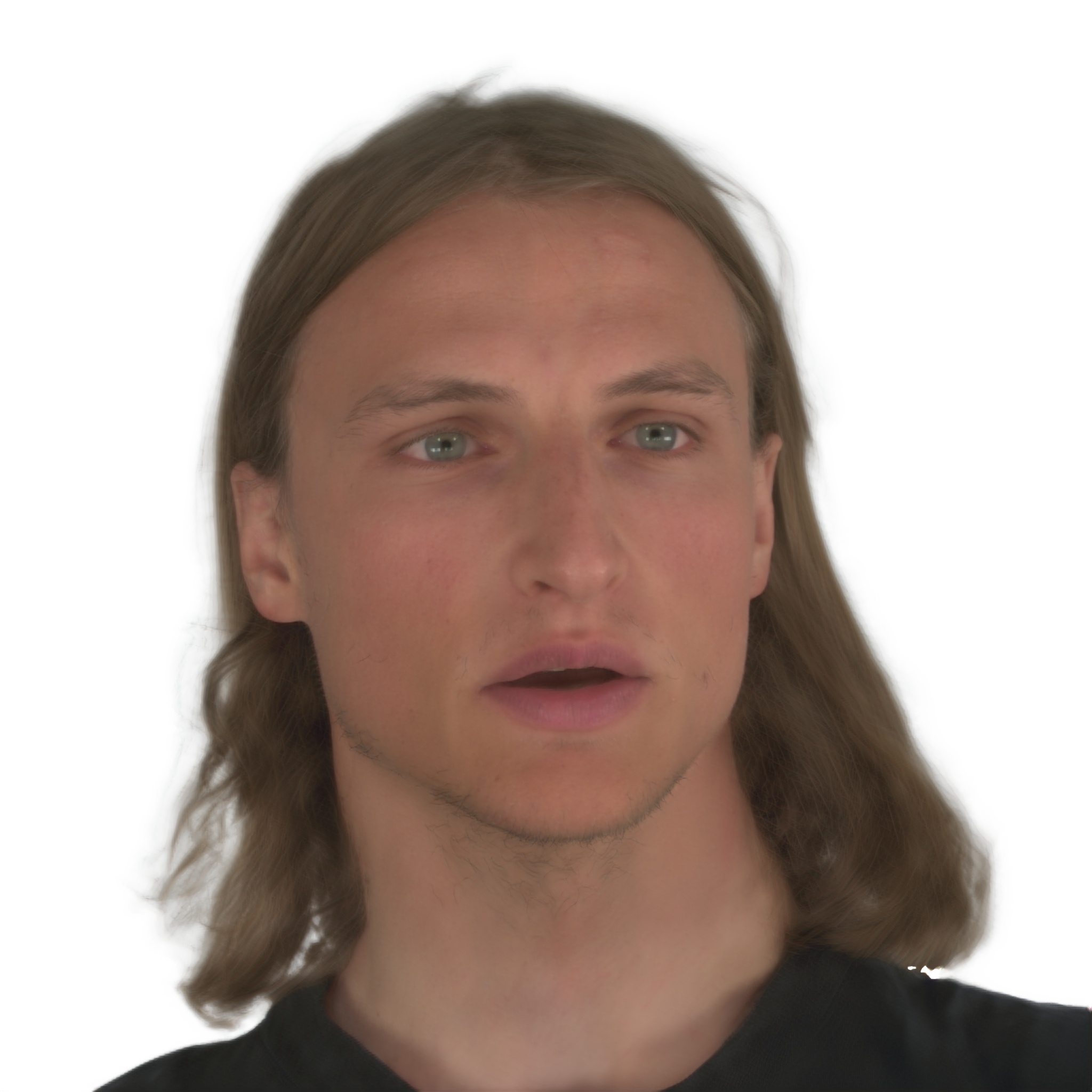}
         \caption{Vanilla}
         \label{fig:example_d445_a443_b}
     \end{subfigure}
     \hfil
     \begin{subfigure}[b]{0.24\textwidth}
         \centering
         \includegraphics[width=\textwidth]{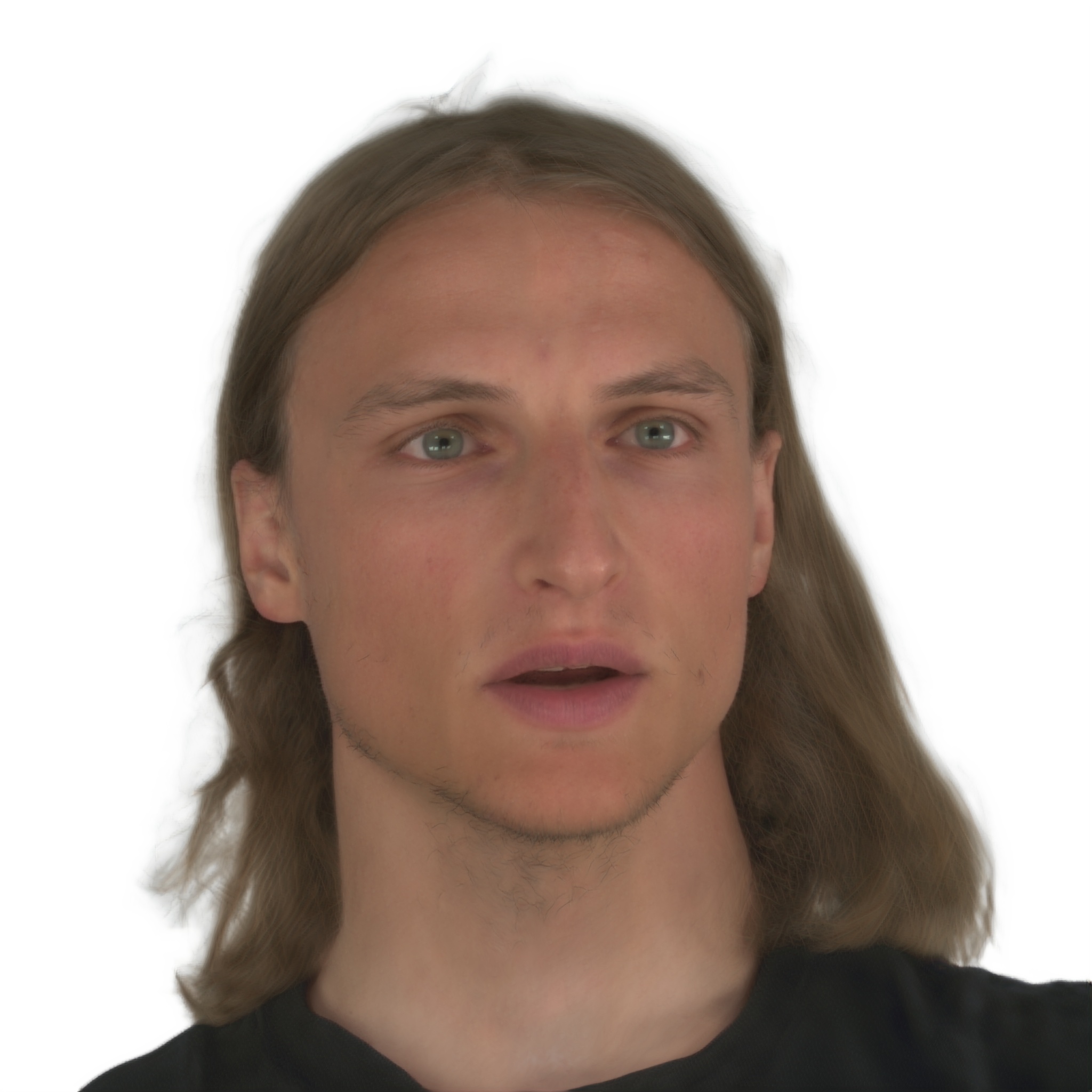}
         \caption{+Random Noise}
         \label{fig:example_d445_a443_c}
     \end{subfigure}
     \hfil
     \begin{subfigure}[b]{0.24\textwidth}
         \centering
         \includegraphics[width=\textwidth]{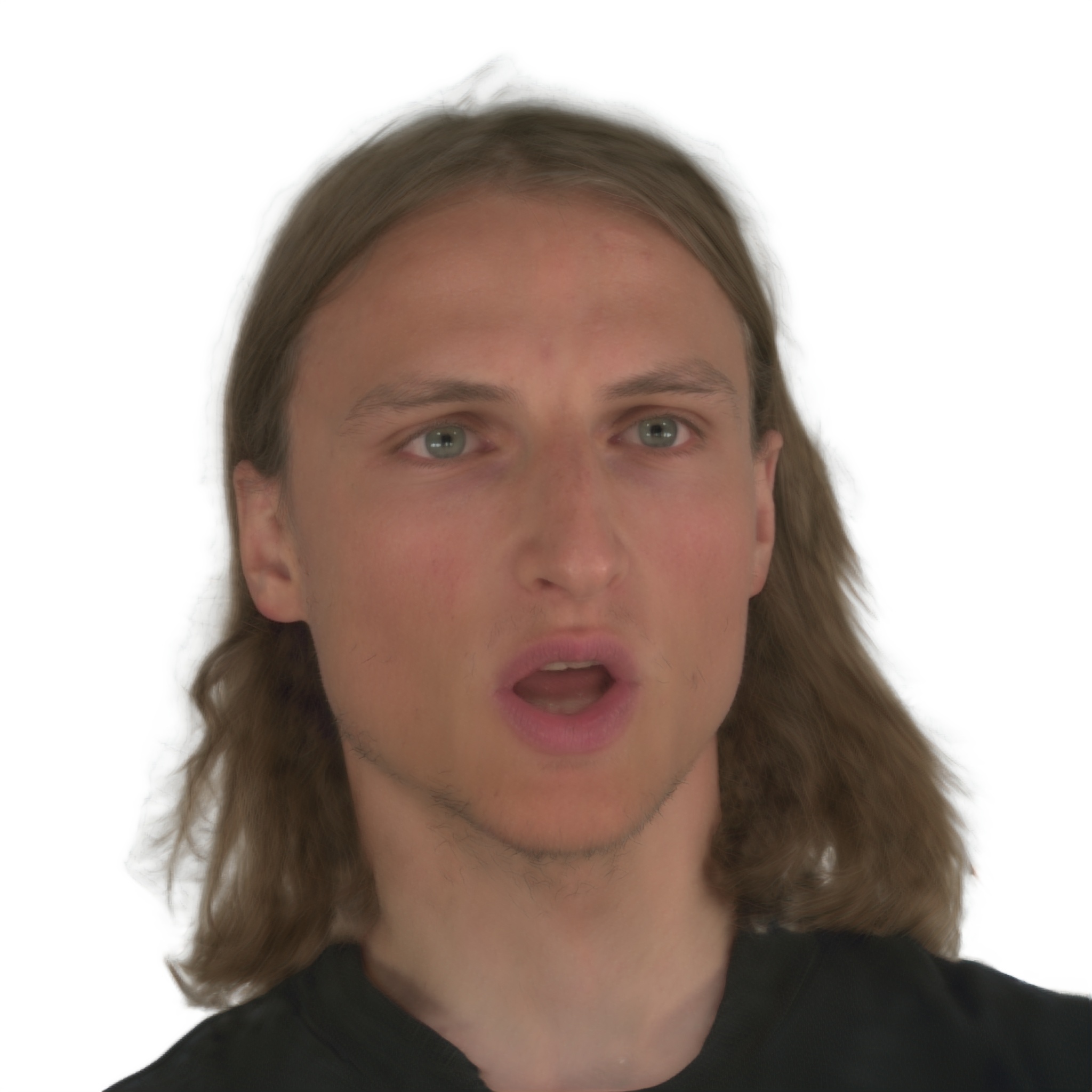}
         \caption{+\ours}
         \label{fig:example_d445_a443_d}
     \end{subfigure}
     \vfil
     \begin{subfigure}[b]{0.24\textwidth}
         \centering
         \includegraphics[width=\textwidth]{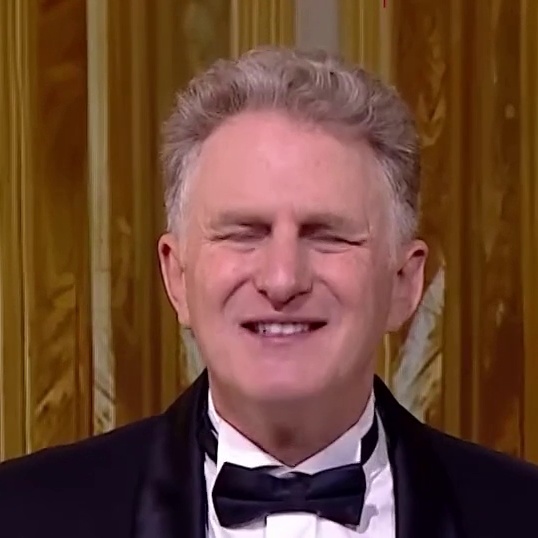}
         \caption{Driver}
         \label{fig:example_Dtrump_A443_a}
     \end{subfigure}
     \hfil
     \begin{subfigure}[b]{0.24\textwidth}
         \centering
         \includegraphics[width=\textwidth]{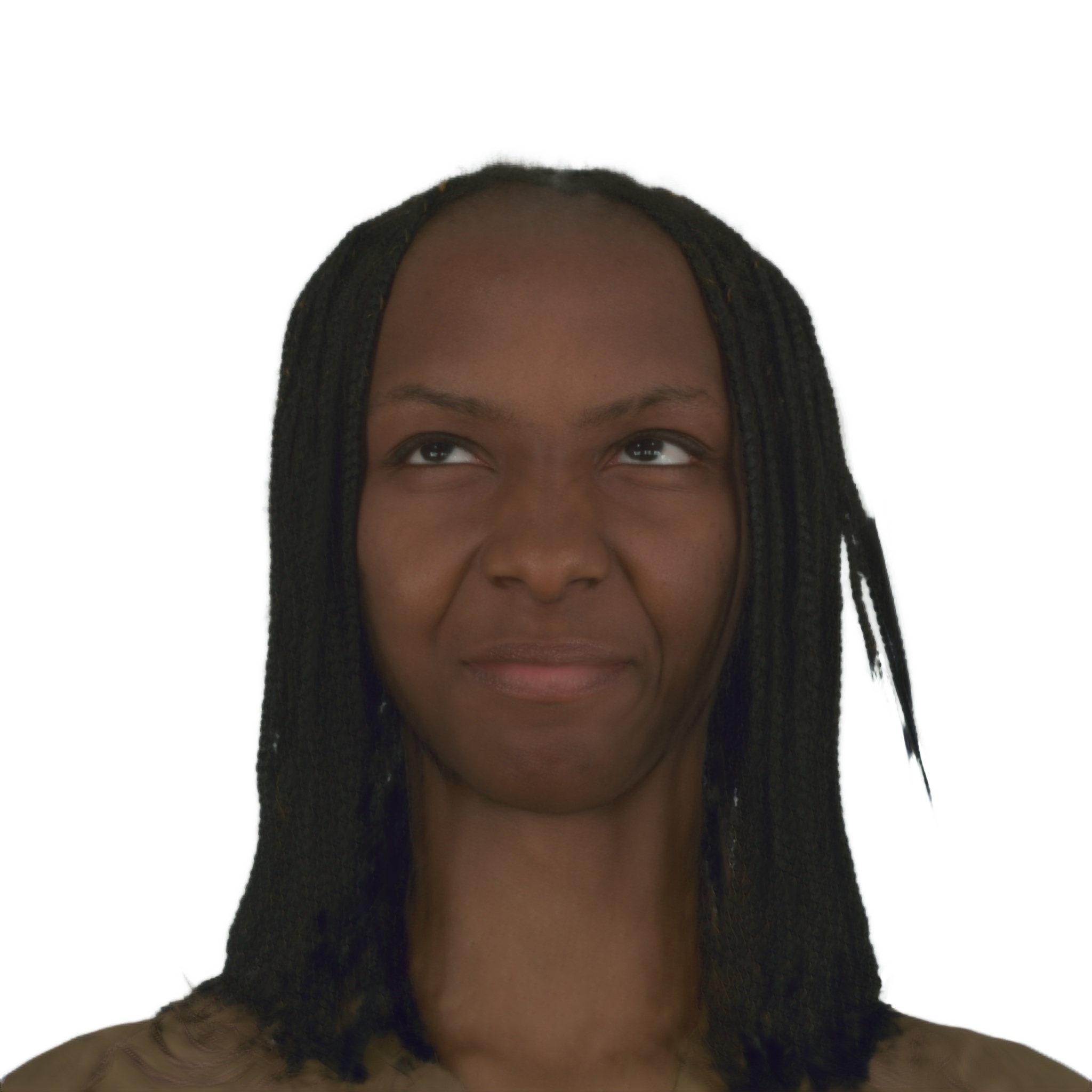}
         \caption{Vanilla}
         \label{fig:example_Dtrump_A443_b}
     \end{subfigure}
     \hfil
     \begin{subfigure}[b]{0.24\textwidth}
         \centering
         \includegraphics[width=\textwidth]{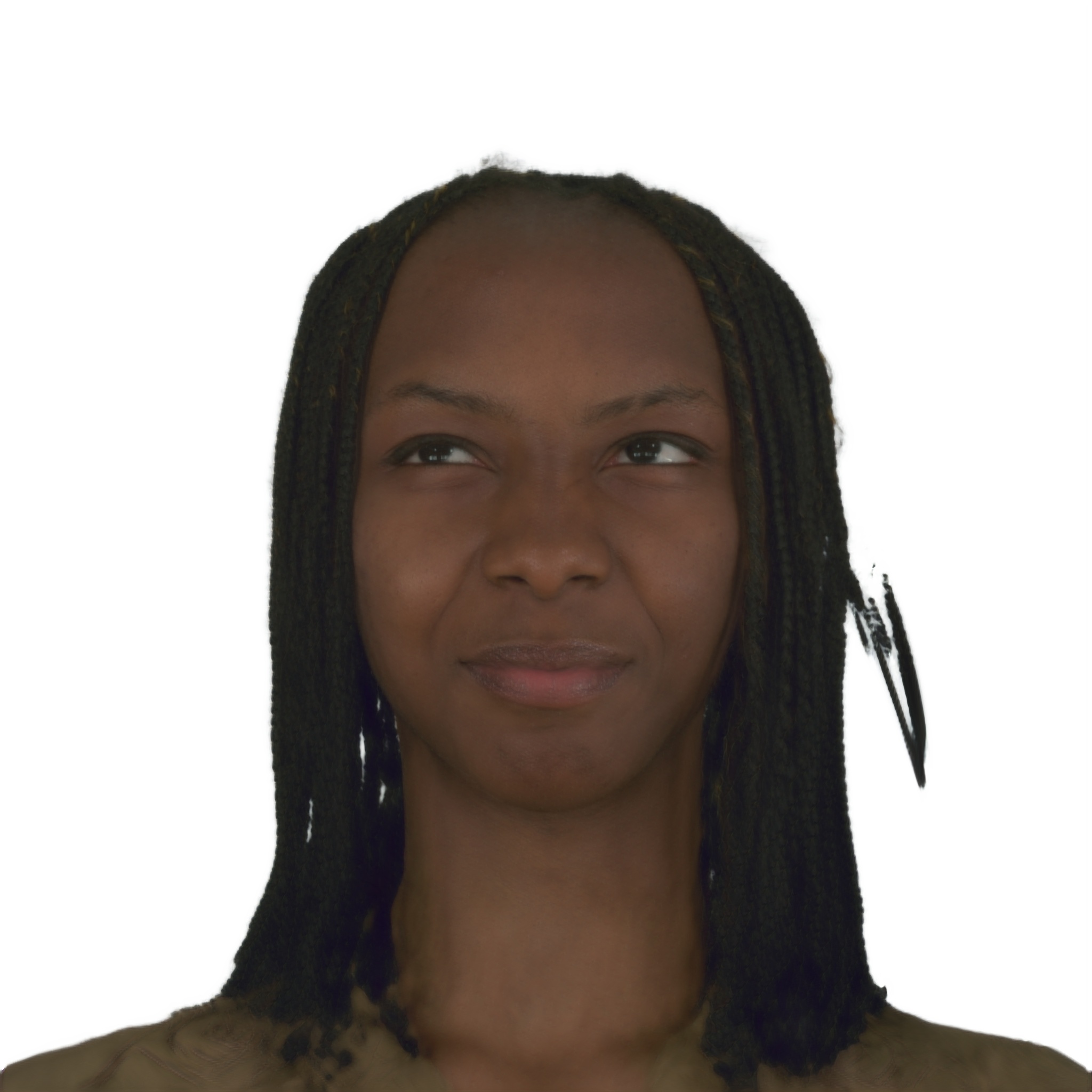}
         \caption{+Random Noise}
         \label{fig:example_Dtrump_A443_c}
     \end{subfigure}
     \hfil
     \begin{subfigure}[b]{0.24\textwidth}
         \centering
         \includegraphics[width=\textwidth]{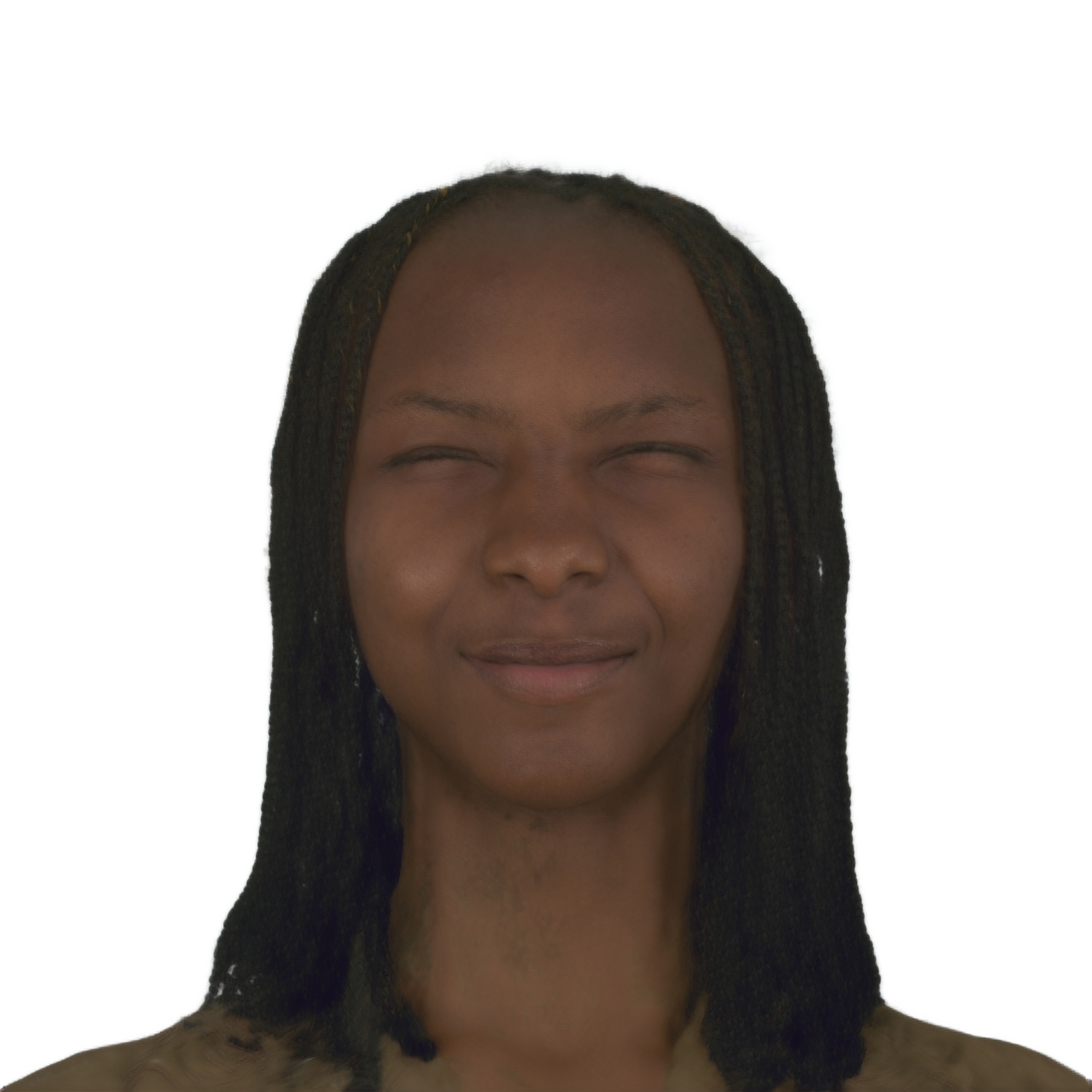}
         \caption{+\ours}
         \label{fig:example_Dtrump_A443_d}
     \end{subfigure}
     \begin{subfigure}[b]{0.24\textwidth}
         \centering
         \includegraphics[width=\textwidth]{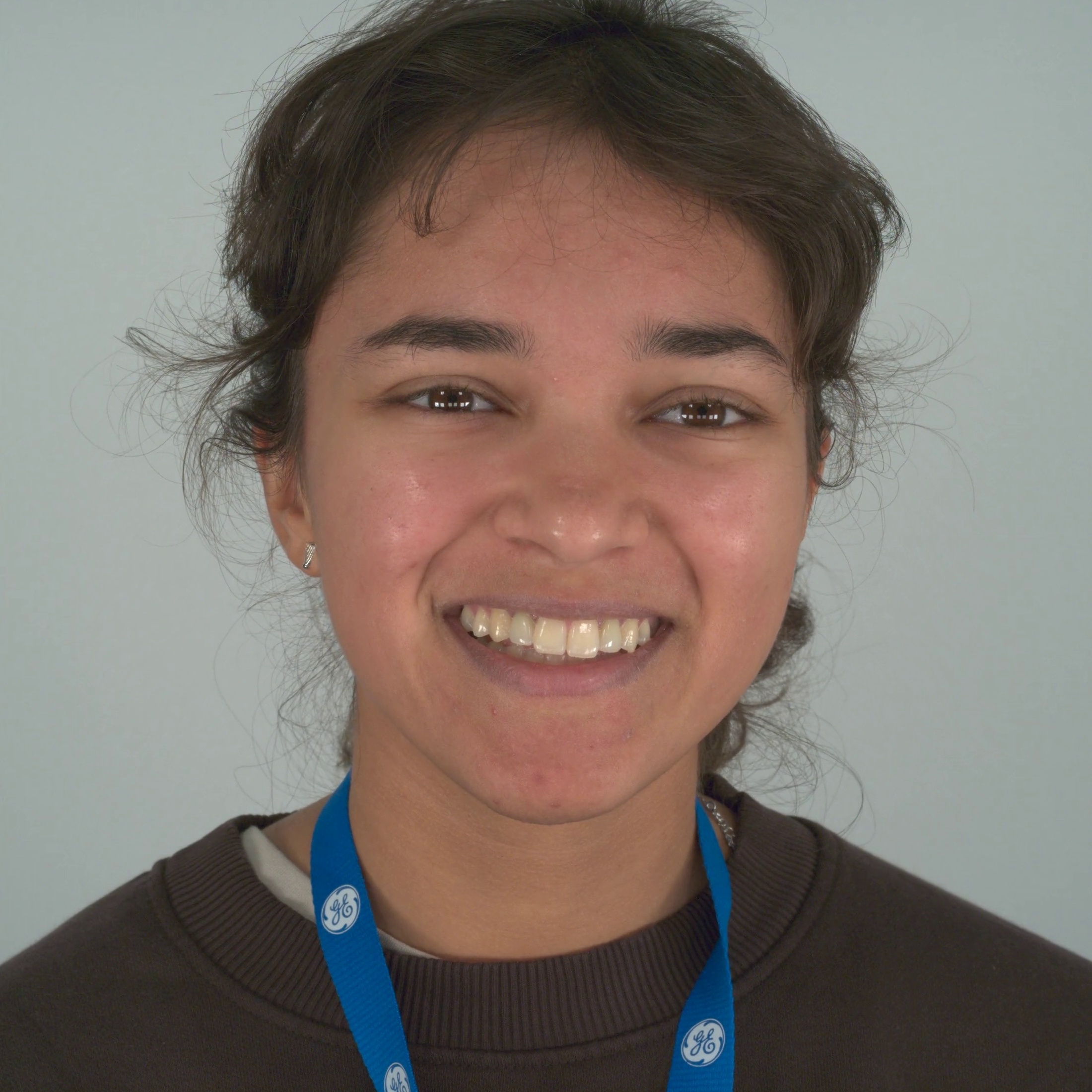}
         \caption{Driver}
         \label{fig:example_Dnvidia_A475_a}
     \end{subfigure}
     \hfil
     \begin{subfigure}[b]{0.24\textwidth}
         \centering
         \includegraphics[width=\textwidth]{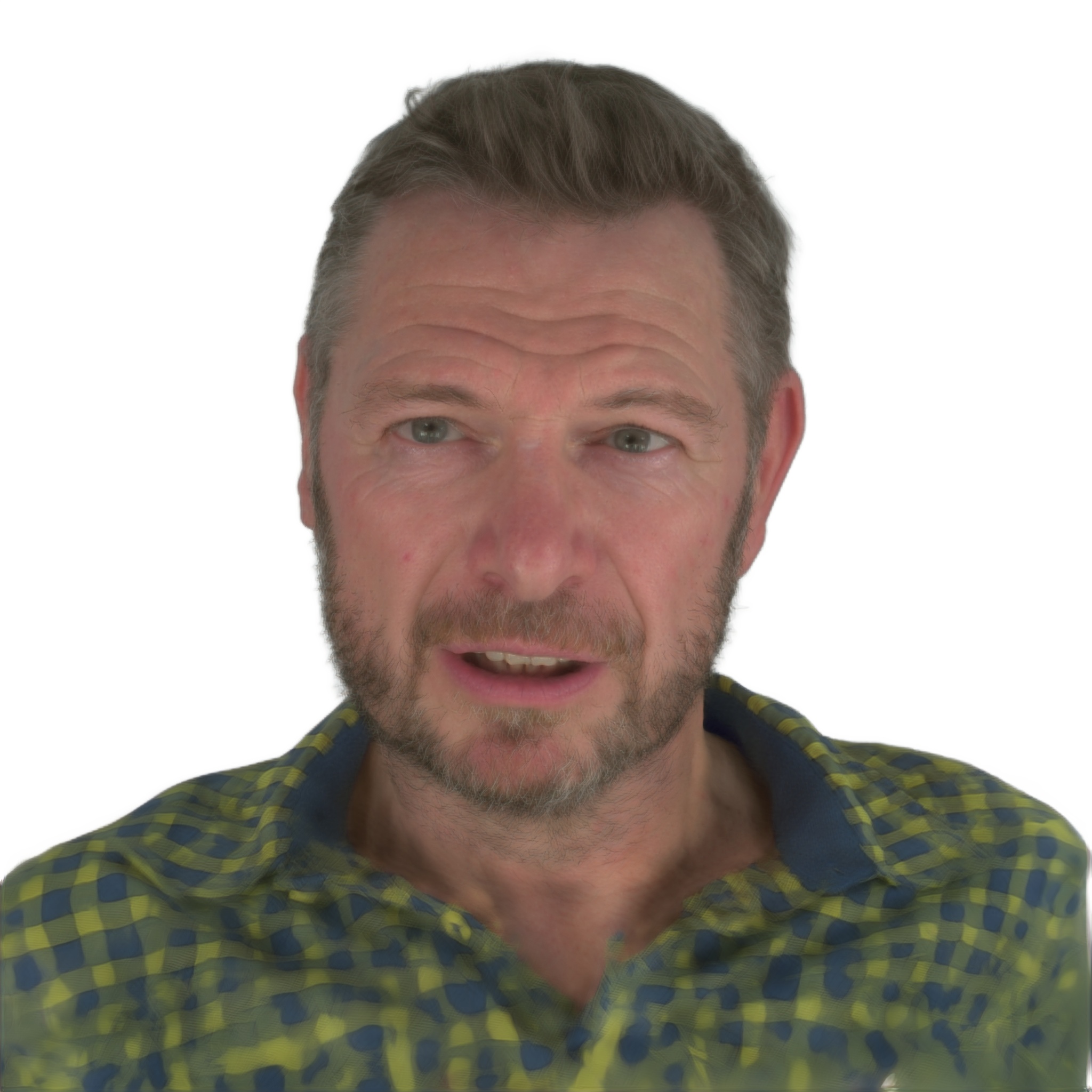}
         \caption{Vanilla}
         \label{fig:example_Dnvidia_A475_b}
     \end{subfigure}
     \hfil
     \begin{subfigure}[b]{0.24\textwidth}
         \centering
         \includegraphics[width=\textwidth]{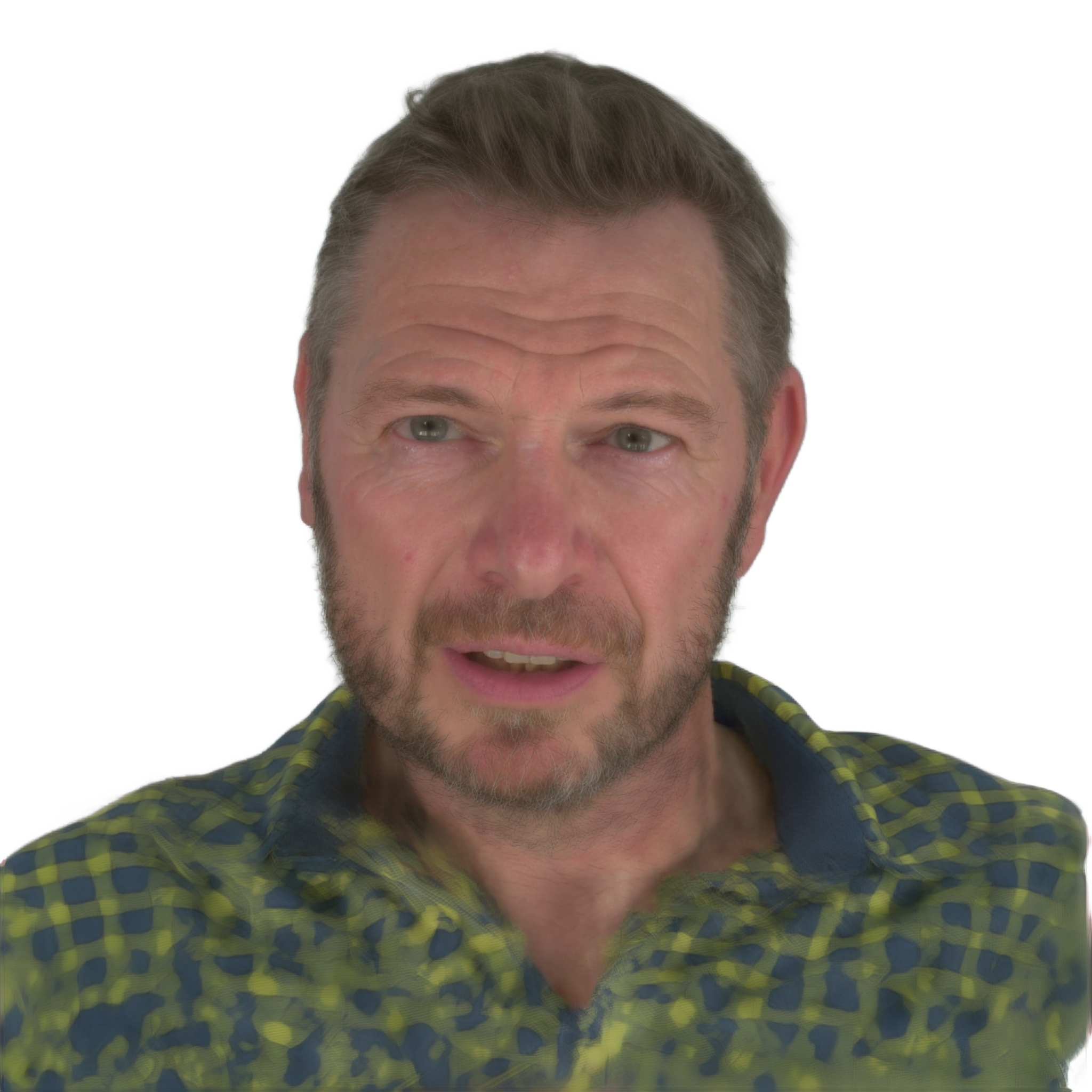}
         \caption{+Random Noise}
         \label{fig:example_Dnvidia_A475_c}
     \end{subfigure}
     \hfil
     \begin{subfigure}[b]{0.24\textwidth}
         \centering
         \includegraphics[width=\textwidth]{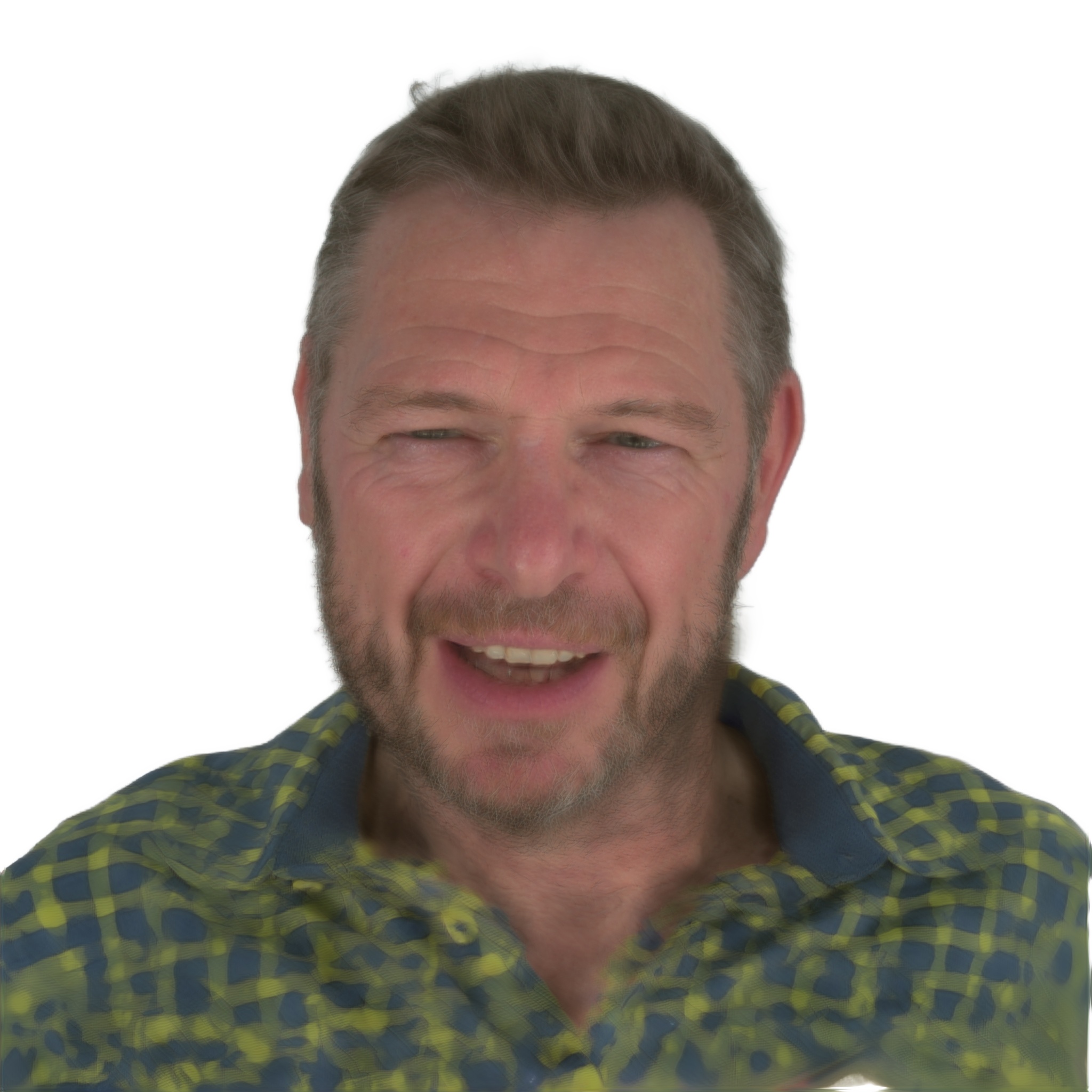}
         \caption{+\ours}
         \label{fig:example_Dnvidia_A475_d}
     \end{subfigure}
     \end{minipage}
}
  \caption[Images]{  
    Cross-identity qualitative comparison. The avatars are taken from the NeRSemble dataset~\cite{kirschstein2023nersemble}. Compared to the baselines, RAF produces expressions that more closely resemble those of the driving subject.} 
\label{fig:example_d445_a443}
\end{figure*}

\section{Introduction}
\label{sec:intro}

Reconstructing a high-fidelity \emph{3D head avatar}, a controllable 3D representation of a human face reconstructed from monocular or multi-view images, has been a long-standing challenge in computer graphics \cite{blanz2023morphable,BFM,booth2018large,FLAME}. Such representations enable photo-realistic rendering of a subject from arbitrary viewpoints and support a wide range of applications, including virtual and augmented reality, telepresence, and digital humans.

Existing head-avatar pipelines broadly follow two learning paradigms. A recent line of work trains large \emph{feed-forward} models \cite{he2025lam,kirschstein2025avat3r} that reconstruct an avatar from a small set of input images, emphasizing fast personalization and convenient capture, but often sacrificing subject-specific fidelity. In contrast, another line of work reconstructs avatars from monocular or sparse-view videos of a single subject, optimizing a subject-specific representation to closely reproduce the observed frames and typically achieving higher visual fidelity and identity preservation
\cite{ultra_high_fid,cap4d,point_avatar,monogaussianavatar,monucular_rgb_video,GafniTZN21,avatarMAV}.

A key design choice in subject-specific avatars is \emph{how facial deformation is modeled}. Historically, many animatable head avatars rely on parametric face models, such as 3D Morphable Models (3DMMs) or FLAME \cite{li2017learning,blanz2023morphable,uniface,face2face,deep_video_portaits}. In these approaches, facial motion is not learned from data but instead inherited from a predefined low-dimensional template space (e.g., blendshapes) and applied via Linear Blend Skinning (LBS). Several recent point-based and 3D Gaussian Splatting (3DGS) avatar systems follow the same paradigm, attaching Gaussians to a tracked mesh and animating them by propagating the template deformation \cite{SEGA,cap4d,gag_avatar,monogaussianavatar,point_avatar}.

While such template-based approaches provide convenient structure and stable control signals, they fundamentally restrict motion to the representational capacity of the template. In practice, this can limit the ability to reproduce \emph{complex, non-linear, or exaggerated} facial motion when the target expression lies outside the template space.

To overcome this limitation, recent work by Xu \etal~\cite{ultra_high_fid} removes the explicit 3DMM scaffold and instead learns expression-conditioned deformation directly for a subject-specific 3D Gaussian avatar. By learning deformation networks rather than inheriting motion from a template, such \emph{template-free} avatars can capture richer facial motion and extend controllability beyond the face region. To the best of our knowledge, it is the state-of-the-art animatable 3DGS head-avatar method that learns expression-dependent deformation network, in a template-free manner, rather than inheriting deformation from a tracked template.

However, removing the template prior introduces a new challenge: \emph{limited expression coverage}. Unlike 3DMM expression spaces that are learned from large multi-identity datasets, template-free deformation networks are typically trained using only a single subject’s capture sequence. Consequently, the deformation model is exposed to only a limited set of expressions during training. Rare or unseen expressions may therefore be poorly modeled, and cross-identity driving can become brittle when the driving motion deviates from the subject’s observed expression distribution.

This highlights a fundamental trade-off. While template-free avatars remove the constraints of fixed template deformation spaces, they also lose the large-scale expression prior that templates implicitly provide. As a result, their performance becomes limited by the expression coverage available in the subject’s training capture.

At the same time, this limitation suggests an opportunity. Since template-free avatars \emph{learn} deformation from data, they can potentially benefit from richer motion supervision during training, provided such supervision can be obtained without requiring additional labeled data or architectural changes.

In this work, we introduce \textbf{RAF} (\emph{Retrieval-Augmented Faces}), a simple training-time strategy for expanding the expression supervision available to template-free Gaussian head avatars. Rather than relying solely on the limited expressions observed in a single subject’s capture, RAF leverages a large multi-identity expression bank to provide additional motion supervision during training. 
Specifically, for a subset of training iterations we replace the subject’s expression features with nearest-neighbor expressions retrieved from this bank while still supervising reconstruction of the subject’s original frames. This encourages the deformation network to explain the subject’s appearance under a broader range of expression conditions, promoting stronger identity-expression disentanglement and improving robustness to expression distribution shift without requiring architectural changes, additional annotations, or paired cross-identity data.

To support this premise, we investigate the effect of retrieval augmentation on the distribution of training expressions. Our analysis indicates that RAF broadens expression diversity and improves coverage of unseen, held-out driver expressions. We further validate the quality of the retrieved neighbors through a perceptual user study demonstrating that nearest-neighbor matches are consistently preferred over alternative matches in both expression and head pose.

Our approach improves both cross-driving and self-driving performance, as demonstrated in \Cref{fig:example_d445_a443}, without requiring architectural modifications, paired cross-identity supervision, or additional annotations. These results suggest that expanding expression coverage during training is a key factor for improving the robustness of template-free avatar models.

Our contributions are summarized as follows:

\begin{itemize}
    \item We introduce \emph{RAF} (Retrieval-Augmented Faces), a simple training-time augmentation that expands the expression supervision available to template-free Gaussian head avatars by substituting expression features with nearest neighbors retrieved from a large unlabeled expression bank.
    
    \item We demonstrate that RAF improves both self-driving and cross-driving performance, yielding more accurate expression reproduction and higher emotion similarity.

    \item We provide empirical analysis showing that retrieval augmentation increases expression diversity and improves coverage of the training distribution, and we validate the perceptual quality of retrieved neighbors through a user study.
\end{itemize}

\section{Related Work}
\label{sec:related_work}

\paragraph{3D Morphable Models (3DMM).}
Early approaches to controllable face modeling, such as the Basel Face Model (BFM)~\cite{BFM} and FLAME~\cite{FLAME}, learn a low-dimensional 3D head representation from scans of many individuals. FLAME in particular has become the de facto backbone for expression control and facial reenactment, combining an identity shape space with articulated components for the jaw, neck, and eyeballs.

A 3DMM encodes geometry in a linear subspace parameterized by identity, expression, and pose coefficients. Fitting a 3DMM to a specific subject is typically performed via optimization over multiple images: identity parameters are shared across all views, while expression and pose coefficients vary per frame. Ideally, expression coefficients should be disentangled from pose. In practice, however, expression and pose dimensions are often entangled, as we demonstrate in \Cref{sec:motivation}. 

Despite these limitations, 3DMMs remain widely used, not only as expression feature extractors but also as structural priors for animatable 3D head avatars. Using a parametric mesh alleviates the burden of geometric modeling, yet constrains the representation to the fixed topology and deformation capacity of the model. As a result, regions such as shoulders or upper garments are typically excluded, and complex non-linear motions (\eg tongue movement) remain out of scope. Although recent updates introduce additional degrees of freedom (\eg explicit eyeball articulation), 3DMMs remain a strong but restrictive inductive bias.  

\paragraph{Gaussian Head Avatars.}
The emergence of neural volumetric representations, including NeRFs \cite{mildenhall2021nerf} and 3D Gaussian Splatting (3DGS) \cite{kerbl20233d}, has shifted head avatar modeling away from meshes toward more flexible radiance-field or point-based primitives. Approaches such as \emph{Gaussian Avatars}~\cite{GaussianAvatars} and \emph{INSTA}~\cite{INSTA} attach 3D Gaussians to a tracked mesh and optimize them for fast, high-fidelity rendering, but inherit their deformation from the tracked mesh. 

Recent animatable Gaussian avatar methods~\cite{GAG, he2025lam, kirschstein2025avat3r} integrate 3DMM priors into a feed-forward reconstruction pipeline. Given a few posed images, a neural network predicts both 3DMM expression parameters and a corresponding 3D Gaussian field, enabling rapid avatar creation at inference time. These approaches leverage large datasets of multiple facial videos to learn generalizable priors. However, their reliance on a specific FLAME~\cite{FLAME} version or similar low-dimensional 3DMMs inherently limits geometric expressiveness and may reduce fidelity for subjects whose facial structure deviates from the template.

Motivated by these limitations, recent work~\cite{ultra_high_fid} removes any fixed 3DMM template and instead learns a \emph{subject-specific} animatable 3DGS by optimizing a canonical (neutral) Gaussian set together with \emph{expression/pose conditioned deformation MLPs}. Concretely, each Gaussian $i$ is represented in a canonical space by a center $\mathbf{x}_i^0$ and learned feature $\mathbf{g}_i$ (and canonical attributes such as rotation/scale/opacity). Given driving signals for expression $\mathbf{e}$ and head pose $\mathbf{p}$, the method predicts residual, motion-dependent updates via separate small MLPs $f_i$ for different factors (\eg expression vs. pose) and different quantities (\eg geometry vs. appearance). In high-level view:
\[
\Delta\mathbf{x}_i^{\text{exp}} = f_{\text{exp}}(\mathbf{g}_i,\mathbf{e}),\qquad
\Delta\mathbf{x}_i^{\text{pose}} = f_{\text{pose}}(\mathbf{g}_i,\mathbf{p}),
\]
and analogously for other Gaussian attributes (rotation/scale/opacity/color). To decouple non-rigid facial motion from more rigid head/neck motion, they blend expression/pose induced updates using spatial weights derived from the distance of each Gaussian to tracked 3D facial landmarks. The deformations are then applied as $\mathbf{x}_i = \mathbf{x}_i^0 + w_i\,\Delta\mathbf{x}_i^{\text{exp}} + (1-w_i)\,\Delta\mathbf{x}_i^{\text{pose}}$. In addition, they propose a geometry-guided initialization that first trains an implicit geometry/deformation model (SDF-based) and transfers it to initialize the Gaussian avatar, improving stability and fidelity. While such template-free avatars avoid the restrictive deformation space and topology of FLAME/3DMMs, our method strengthens them by explicitly addressing cross-identity expression transfer fidelity.

A key distinction among Gaussian-head-avatar methods lies in their training requirements and generalization behavior. Some approaches achieve high fidelity by training on subject-specific multi-view captures~\cite{ultra_high_fid,INSTA,GaussianAvatars}, whereas others aim for few-shot or even single-image reconstruction~\cite{kirschstein2025avat3r,chu2024generalizable,he2025lam}, typically relying on FLAME as a geometric prior. These methods balance reconstruction quality, controllability, and data efficiency depending on how strongly they depend on structural priors. In parallel, Zielonka \etal~\cite{zielonka2025gaussian} explore improving efficiency, such as Gaussian Eigen Models (GEM)~\cite{zielonka2025gaussian}, which compress an optimized Gaussian avatar into a low-rank basis for real-time reenactment. 

However, the prior pipelines obtain controllability through a fixed deformation parameterization, typically FLAME/3DMM blendshapes and skinning, where the expression space is specified a priori rather than learned from data. This design choice fundamentally differs from template-free avatars that learn expression-conditioned deformation fields, and it also makes training-time augmentation strategies like \ours less applicable: \ours is specifically designed to improve learned deformation generalization by expanding the effective expression supervision for the benefit of deformation MLPs.

\section{Motivation}
\label{sec:motivation}
Our RAF approach relies on a simple premise: expression features provide an identity-agnostic notion of similarity that enables reliable cross-identity matching. We explicitly target improving cross-driving where, expressions from diverse source subjects (drivers) must transfer faithfully to the target identity. A natural concern is that nearest-neighbor retrieval could either be \emph{too far} - introducing mismatched expressions and acting as ``noise'', or, \emph{too close} - adding little new information beyond what the subject already exhibits. In this section, we show that neither failure mode dominates in practice: NN retrieval applied during training meaningfully increases effective expression diversity and coverage, reducing the gap between expressions seen during training and those encountered at test time. A related subtlety we later demonstrate is that expression embeddings are often entangled to head pose: while the same expression can occur under different poses, nearby points in expression space frequently share similar pose cues.

\begin{wraptable}{r}{0.50\linewidth}
\vspace{-2.0\baselineskip} 
\centering
\small
\caption{Coverage analysis of expression features. Applying RAF substantially improves coverage of the test expression distribution.}
\label{tab:bank_analysis}
\resizebox{0.5\columnwidth}{!}{%
\begin{tabular}{lcccccc}
\toprule
\multirow{2}{*}{Subject} &
\multicolumn{2}{c}{MMD $\downarrow$} &
\multicolumn{2}{c}{KL $\downarrow$} &
\multicolumn{2}{c}{B2T Dist $\downarrow$} \\
\cmidrule(lr){2-3} \cmidrule(lr){4-5} \cmidrule(lr){6-7}
& Vanilla & +RAF & Vanilla & +RAF & Vanilla & +RAF \\
\midrule
388 & 0.254 & \textbf{0.216} & 38.948 & \textbf{25.365} & 1.398 & \textbf{0.796} \\
422 & 0.186 & \textbf{0.128} & 50.575 & \textbf{31.425} & 1.001 & \textbf{0.560} \\
443 & 0.131 & \textbf{0.103} & 33.078 & \textbf{19.658} & 0.763 & \textbf{0.533} \\
445 & 0.152 & \textbf{0.089} & 56.310 & \textbf{32.658} & 0.872 & \textbf{0.499} \\
475 & 0.213 & \textbf{0.167} & 35.662 & \textbf{21.417} & 1.023 & \textbf{0.583} \\
\midrule
Average & 0.187 & \textbf{0.141} & 42.915 & \textbf{26.105} & 1.011 & \textbf{0.594} \\
\bottomrule
\end{tabular}
}
\vspace{-1.8\baselineskip} 
\end{wraptable}

\Cref{tab:bank_analysis} presents how well each subject’s \emph{training} expression feature distribution matches the unseen \emph{drivers} expression distribution, before and after applying retrieval augmentation. To ensure a controlled comparison, both distributions contain the same number of samples $N$: the \textit{Vanilla} set consists of $N$ expression features drawn from the subject’s training frames, whereas \textit{+RAF} uses $N$ features formed by mixing $\tfrac{N}{2}$ subject training expressions with $\tfrac{N}{2}$ nearest-neighbor expressions retrieved from the expression bank (for those sampled training frames). Although +RAF replaces half of the subject’s own training expressions, it yields a closer match to the multi-driver test distribution, indicating improved coverage of unseen expression diversity. 

We report three metrics. Maximum Mean Discrepancy (MMD) and KL divergence quantify how close a subject's training expression distribution is to the test set. Bank-to-Train (B2T) distance measures coverage: for each test expression, it reports the average distance to its nearest training neighbor, capturing how accessible unseen expressions are given the avatar's training data. Across all subjects, +RAF consistently reduces MMD, KL, and B2T, by $25\%$, $39\%$ and $41\%$ respectively, indicating that retrieval augmentation improves coverage of the subject’s expression diversity. 
\begin{figure*}[ht]
    \centering
    \includesvg[width=0.8\linewidth]{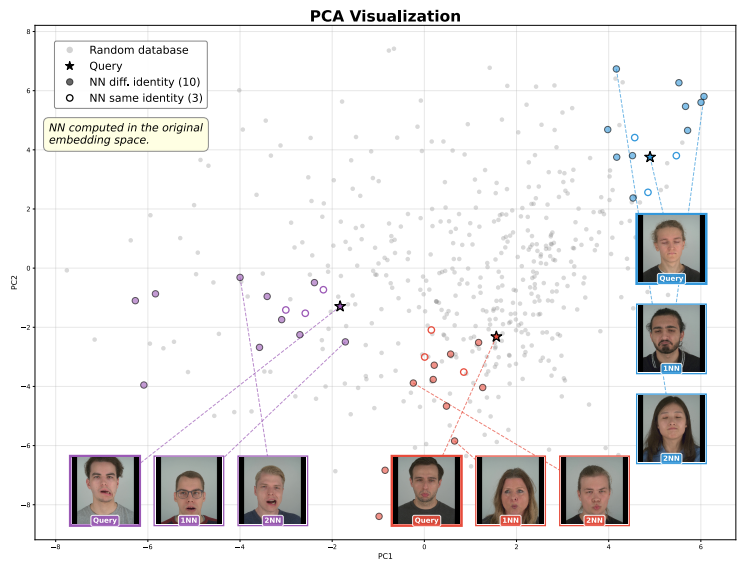}
    \caption{PCA visualization of the expression bank. We project expression features into 2D with PCA. For three query expressions (stars), we highlight their $10$ nearest neighbors computed in the original BFM expression-feature space. Despite the projection, each query’s neighbors form a cluster of similar expressions, even though they belong to different identities. This indicates that the expression bank contains meaningful cross-identity matches, which enables reliable nearest-neighbor retrieval for expression substitution during training.
}
    \label{fig:pca_nn}
\end{figure*}

Next, we visualize the structure of our expression bank in \Cref{fig:pca_nn}, using a PCA projection. For three different query expressions (stars), we plot their $10$ nearest neighbors (share the same color) in the expression space. Although PCA preserves global variation, it does not guarantee alignment of fine-grained expression clusters; nonetheless, each query’s nearest neighbors form tight, semantically coherent groups, despite belonging to different identities.  
This demonstrates that the expression bank contains meaningful cross-identity matches that are locally similar in the original feature space, validating the use of nearest-neighbor retrieval as an effective mechanism for providing cross-identity expression supervision.

Finally, we conduct a user study to verify that nearest neighbors in our expression feature space are also perceptually similar to human observers. We use Amazon Mechanical Turk (AMT) and sample $1k$ query images. For each query, we form two candidate matches and ask workers to select which candidate better matches the query in terms of \emph{(i) head pose} and \emph{(ii) facial expression} (different identities are always shown). We evaluate two comparisons: (1) the nearest neighbor (NN) from the full expression bank versus a random bank image, and (2) the NN from the full bank versus the NN retrieved from a much smaller bank. We include the head-pose question both to encourage workers to judge expression similarity beyond pose alignment, and to probe how much pose information is implicitly captured by the expression embedding.

\begin{wrapfigure}{r}{0.5\textwidth}
    \vspace{-1.5\baselineskip} 
    \centering
    \includegraphics[width=\linewidth]{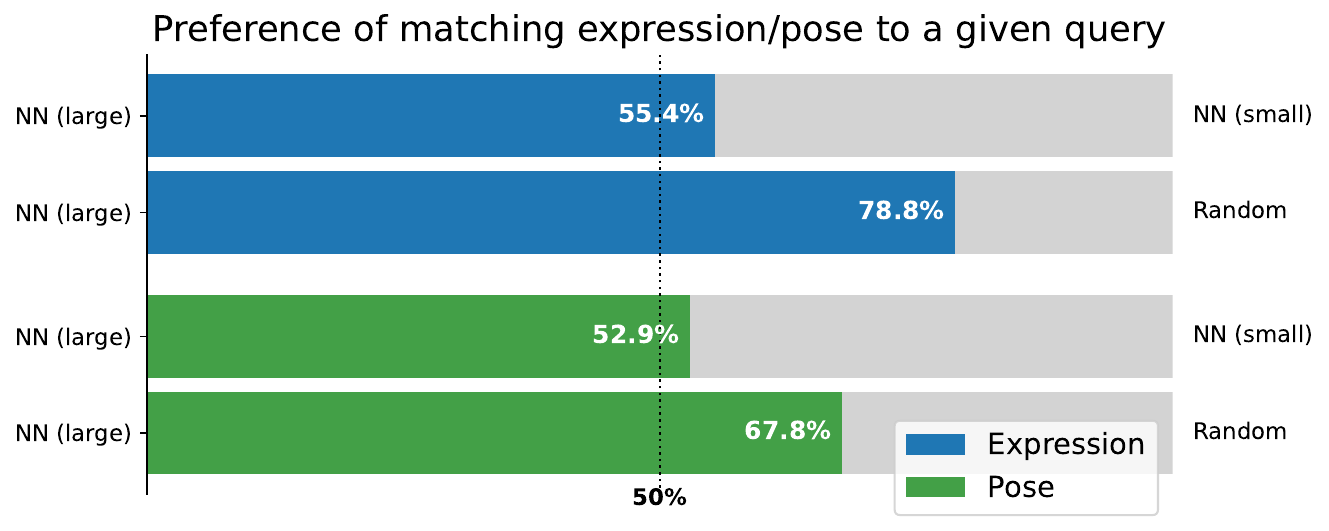}
    \caption{Human preference of most similar match to a query face, reported separately for head pose and facial expression. Nearest neighbors (NN) retrieved from a large/small expression bank, built from the same set of identities but with fewer frames.}
    \label{fig:amt}
    \vspace{-1.8\baselineskip} 
\end{wrapfigure}

\Cref{fig:amt} reports preference rates. Across both head pose and facial expression, workers strongly prefer nearest-neighbor (NN) matches over random ones,
indicating that retrieval produces perceptually meaningful correspondences rather than arbitrary pairs. 
This finding also mitigates the concern that nearest neighbors in a high-dimensional space may not correspond to perceptually similar faces: human observers readily distinguish retrieved matches from random alternatives. Notably, pose is also preferred for NN retrieval despite using \emph{only} expression features for matching, suggesting that the expression embedding is entangled with head-pose cues in this feature space - an observation we later validate in our experiments. We also observe a consistent, though smaller, advantage when retrieving from the full bank compared to a smaller bank, supporting the benefit of scale for improving retrieval quality.
\section{Method}
\label{sec:method}
Our objective is to improve the expression controllability of a \emph{subject-specific} Gaussian head avatar reconstructed from that subject’s available monocular video data. In particular, we aim to enhance both \emph{self-driving} (expressions originating from the same identity) and \emph{cross-driving} (expressions originating from a different identity). The key idea is to augment the avatar’s training with expressions drawn from \emph{other} subjects, thereby exposing the model to a much broader and more diverse expression space than the one present in the subject’s own captures.
\begin{figure*}[ht]
	\centering
	\includesvg[width=0.9\linewidth]{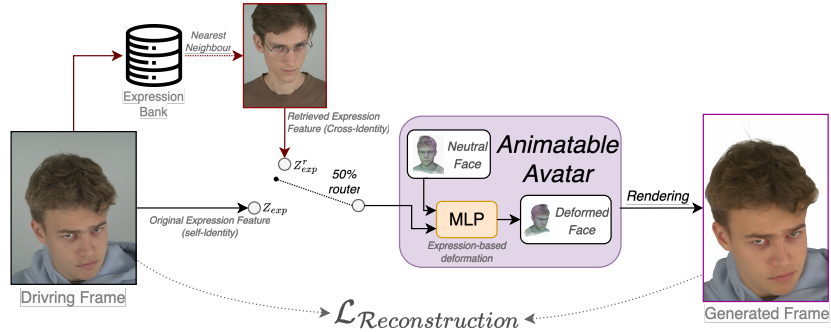}
	\caption{An overview of our method. During training, we randomly replace 50\% of each subject’s expression features with those of their nearest neighbors originating from different identities. This retrieval-based substitution encourages the model to generalize expressions across identities and disentangle expression from appearance. The avatar network (MLP) is subsequently trained to reconstruct the original frame conditioned on the substituted expression features, optimized under a reconstruction objective.}
	\label{fig:method}
\end{figure*}
\paragraph{Method Overview.}
\ours accomplishes this by substituting the subject’s native expression features with semantically matched expression features retrieved from a large multi-identity corpus (Expression Bank). The avatar continues to be supervised using its own ground-truth frames, yet it is conditioned on expressions that may come from entirely different individuals. This disrupts the tight identity-expression coupling in standard training and encourages expression generalization beyond the subject's observed motions. \Cref{fig:method} presents an overview of our method.

\paragraph{Problem Formulation.}

Let a training subject be represented by a set of monocular video frames
\(
\mathcal{D} = \{ I_t \}_{t=1}^{T},
\)
from which a subject-specific Gaussian head avatar is optimized. Following Xu \etal \cite{ultra_high_fid}, the avatar consists of a set of 3D Gaussians $\mathcal{G}$ and a motion field $f_\theta$ that conditions the Gaussians on an \emph{expression feature vector} $\mathbf{e} \in \mathbb{R}^d$.

For each frame $I_t$, BFM provides an estimated expression vector $\mathbf{e}_t$. Standard training conditions the avatar on $\mathbf{e}_t$ and supervises it to reconstruct $I_t$. Let $f_\theta(\mathcal{G}, \mathbf{e}_t)$ denote the expression-conditioned deformation of the Gaussian field, and let $\mathcal{R}(\cdot)$ be the differentiable 3DGS renderer. The reconstruction loss is therefore given by $\mathcal{L}_{\text{self}} = \Sigma_{l\in\{\mathcal{L}_1,\mathcal{L}_{VGG}\}}\lambda_{l}\|\, \mathcal{R}( f_\theta(\mathcal{G}, \mathbf{e}_t) ) - I_t \,\|_{l}$, a weighted sum of the $\mathcal{L}_1$ and the perceptual loss ($\mathcal{L}_{VGG}$).

This setup tightly binds identity and expression: the avatar only observes expressions performed by this specific subject, limiting its ability to follow expression controls from other identities at test time.

\subsection{Retrieval-Augmented Expression Substitution}

The key idea behind \ours is to replace the subject's native expression feature $\mathbf{e}_t$ with a semantically similar expression originating from a \emph{different} identity. This exposes the avatar to expression controls that do not appear in the subject’s own video, broadening the expression space encountered during training.

Let $\mathcal{B} = \{ (\mathbf{e}_i, I_i) \}_{i=1}^{N}$ denote a large expression bank constructed from many external subjects (Sec.~\ref{sec:expression_bank}). For each training frame $I_t$, we retrieve the nearest expression feature in this bank: 
$\hat{\mathbf{e}}_t = \arg\min_{\mathbf{e}_i \in \mathcal{B}} \| \mathbf{e}_i - \mathbf{e}_t \|_2$, subject to $\mathrm{ID}(I_i) \neq \mathrm{ID}(I_t)$. 
The avatar is then conditioned on $\hat{\mathbf{e}}_t$ but still supervised to reconstruct $I_t$, yielding the retrieval-augmented loss: 
$$\mathcal{L}_{\text{\ours}} = \Sigma_{l\in\{\mathcal{L}_1,\mathcal{L}_{VGG}\}}\lambda_{l}\|\, \mathcal{R}( f_\theta(\mathcal{G}, \hat{\mathbf{e}}_t) ) - I_t \,\|_{l}$$

By conditioning on cross-identity expressions while supervising against the original frame, the model learns to apply novel expressions to the subject's appearance, effectively decoupling expression from identity without requiring any paired cross-identity data.

\subsection{Expression Bank}
\label{sec:expression_bank}

To supply diverse expression examples, we construct an expression bank from the NeRSemble dataset~\cite{kirschstein2023nersemble}. We collect approximately $83\text{k}$ frames across $415$ subjects. For each frame, we extract a 3DMM expression feature $\mathbf{e}$ using the BFM tracker. These vectors are stored in a searchable index, forming the set $\mathcal{B}$.

The bank provides broad coverage of expression types, intensities, and identity-specific variations. This diversity is essential for RAF, as it ensures that for most subject expressions $\mathbf{e}_t$, a meaningful cross-identity match $\hat{\mathbf{e}}_t$ can be retrieved. Retrieval examples demonstrate that the nearest neighbors generally share strong expression similarity despite differing identities, as we find in \Cref{sec:motivation}.

\subsection{Mixed Training Strategy}

Directly replacing all expression features with cross-identity matches may cause the avatar to drift away from the subject’s native motion space. To avoid this, RAF employs a mixed strategy: for each training frame, the expression feature is replaced with a retrieved feature $\hat{\mathbf{e}}_t$ with probability $p$ (we use $p=0.5$). Otherwise, the original $\mathbf{e}_t$ is used.

This mixed strategy yields a balanced curriculum: frames that use the native expression features help preserve the subject's authentic motion priors, while frames that use cross-identity features expose the model to a broader expression space and encourage generalization to cross-driving scenarios.

The final training objective is therefore a mixture of self-driving and retrieval-augmented reconstruction losses:
$\mathcal{L} = (1-p)\,\mathcal{L}_{\text{self}} + p\,\mathcal{L}_{\text{RAF}}$.

This simple strategy provides stable optimization while significantly increasing expression diversity and improving both self- and cross-driving fidelity.




\section{Evaluation}
\label{sec:evaluation}
\begin{table*}[ht]
\centering
\small
\caption{Quantitative comparison of our method with the original and random-noise baselines on a test set from the NeRSemble dataset~\cite{kirschstein2023nersemble}. Best results are in bold. RAF outperforms both baselines across most metrics in both self- and cross-driving settings, despite being designed primarily for the cross-driving task. In particular, RAF achieves a substantial improvement in AED under cross-driving, indicating a more accurate reproduction of the driver’s expressions.}
\label{tab:results}
\resizebox{\columnwidth}{!}{%
\begin{tabular}{lcccccccc}
\toprule
\multirow{2}{*}{Method} & 
\multicolumn{5}{c}{Self-Driving} & 
\multicolumn{3}{c}{Cross-Driving} \\ 
\cmidrule(lr){2-6} \cmidrule(lr){7-9}
& $\uparrow$ PSNR & $\uparrow$ SSIM & $\downarrow$ AED & $\downarrow$ APD & $\uparrow$ Emotion Sim. 
& $\downarrow$ AED & $\downarrow$ APD & $\uparrow$ Emotion Sim. \\ 
\midrule
Original & {\bf 18.038} & 0.7881 & 0.24227 & 0.13409 & 0.82016 & 0.27037 & 0.12289 & 0.78746 \\
Random Noise & 18.004 & 0.7890 & 0.24998 & 0.13023 & 0.81604 & 0.26920 & {\bf 0.12105} & 0.78770 \\

Ours (RAF) & 18.002 & {\bf 0.7917} & {\bf 0.23094} & {\bf 0.12932} & {\bf 0.85595} & {\bf 0.26164} & 0.12687 & {\bf 0.80801} \\
\bottomrule
\end{tabular}
  }
  
\end{table*}

In this section, we evaluate the effectiveness of our \ours approach on the task of self- and cross-driving. We adopt the NeRSemble-benchmark, constructed from the NeRSemble dataset~\cite{kirschstein2023nersemble}, consisting of $5$ distinct subjects. For each subject, we train a high-fidelity Gaussian Head Avatar~\cite{ultra_high_fid} using all monocular videos \emph{except} the ``FREE'' sequence. The FREE sequence in NeRSemble is a longer, unconstrained recording in which the participant performs arbitrary expressions and head motions, thus providing the richest and most diverse expression set for evaluation.

\textbf{Training setup: }
We apply our \ours approach to each avatar during training. We compare our approach against two baselines: \emph{Vanilla}: the original underlying avatar method \cite{ultra_high_fid}, \emph{Random Noise}: the expression vector is perturbed with Gaussian noise during training, a regularization strategy shown to be beneficial in related disentanglement  contexts~\cite{gabbay2020lord}. We deliberately compare RAF against the same underlying architecture \cite{ultra_high_fid} with and without our augmentation, rather than against different avatar methods, as RAF is a training-time augmentation and not a new architecture. Comparing a FLAME-based feed-forward method against our template-free baseline would conflate architectural differences with the effect of our augmentation strategy.

\begin{figure*}[ht]
    \centering
    \resizebox{0.7\textwidth}{!}{
    \begin{minipage}{\textwidth}
    
    \begin{subfigure}[b]{0.24\columnwidth}
         \centering
         \includegraphics[width=\textwidth]{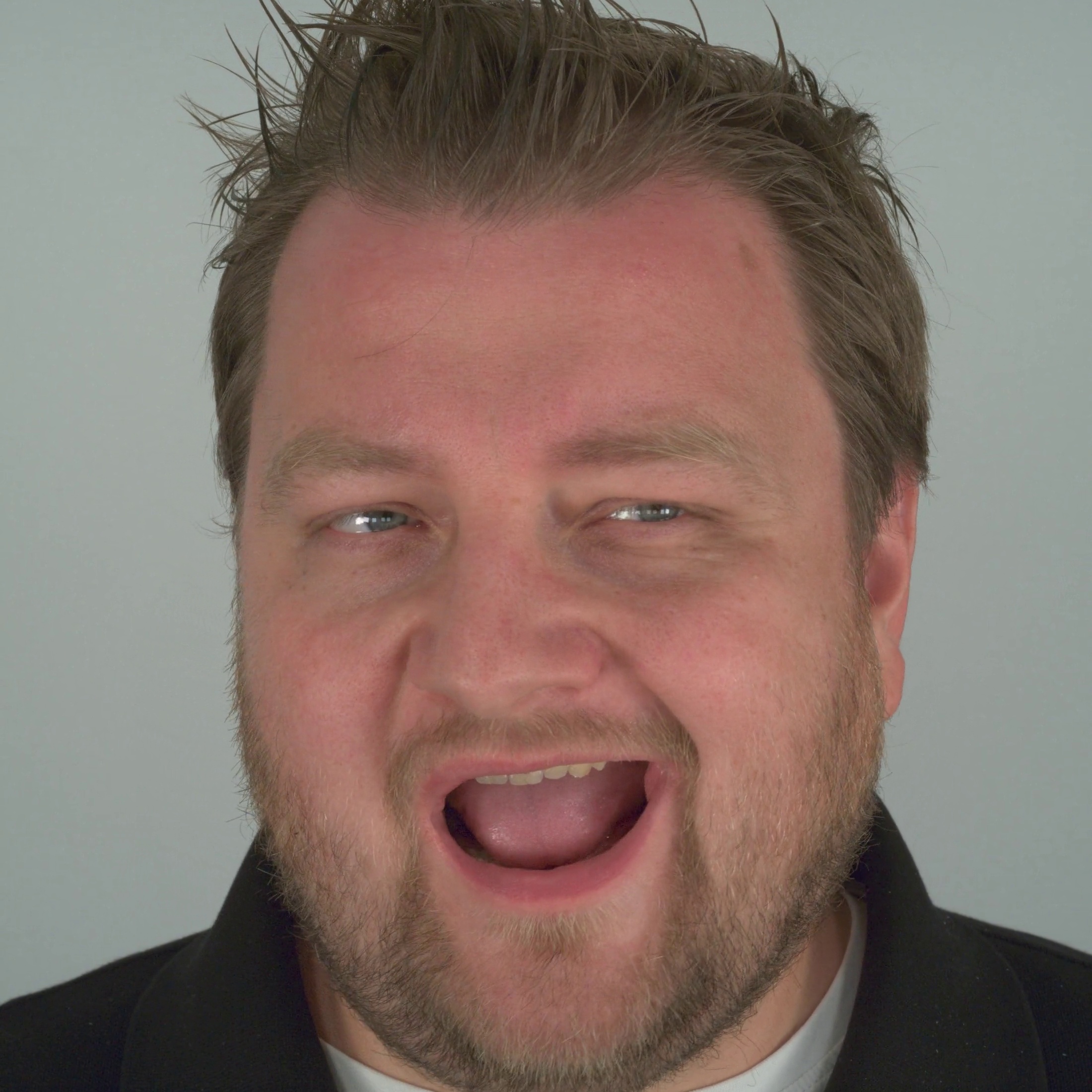}
         \caption{Driver}
         \label{fig:self_388_driver}
     \end{subfigure}
     \hfil
     \begin{subfigure}[b]{0.24\columnwidth}
         \centering
         \includegraphics[width=\textwidth]{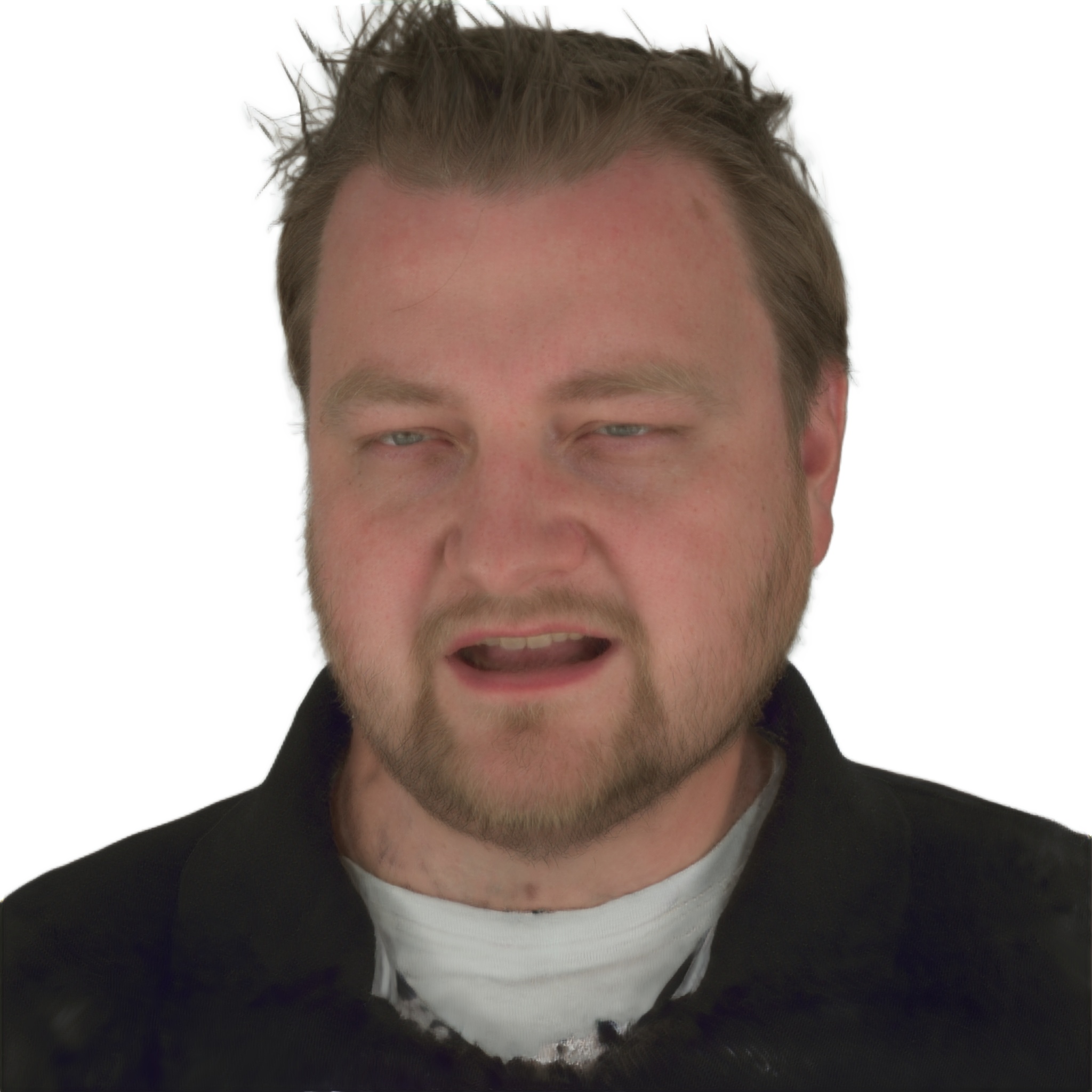}
         \caption{Original}
         \label{fig:self_388_original}
     \end{subfigure}
     \hfil
     \begin{subfigure}[b]{0.24\columnwidth}
         \centering
         \includegraphics[width=\textwidth]{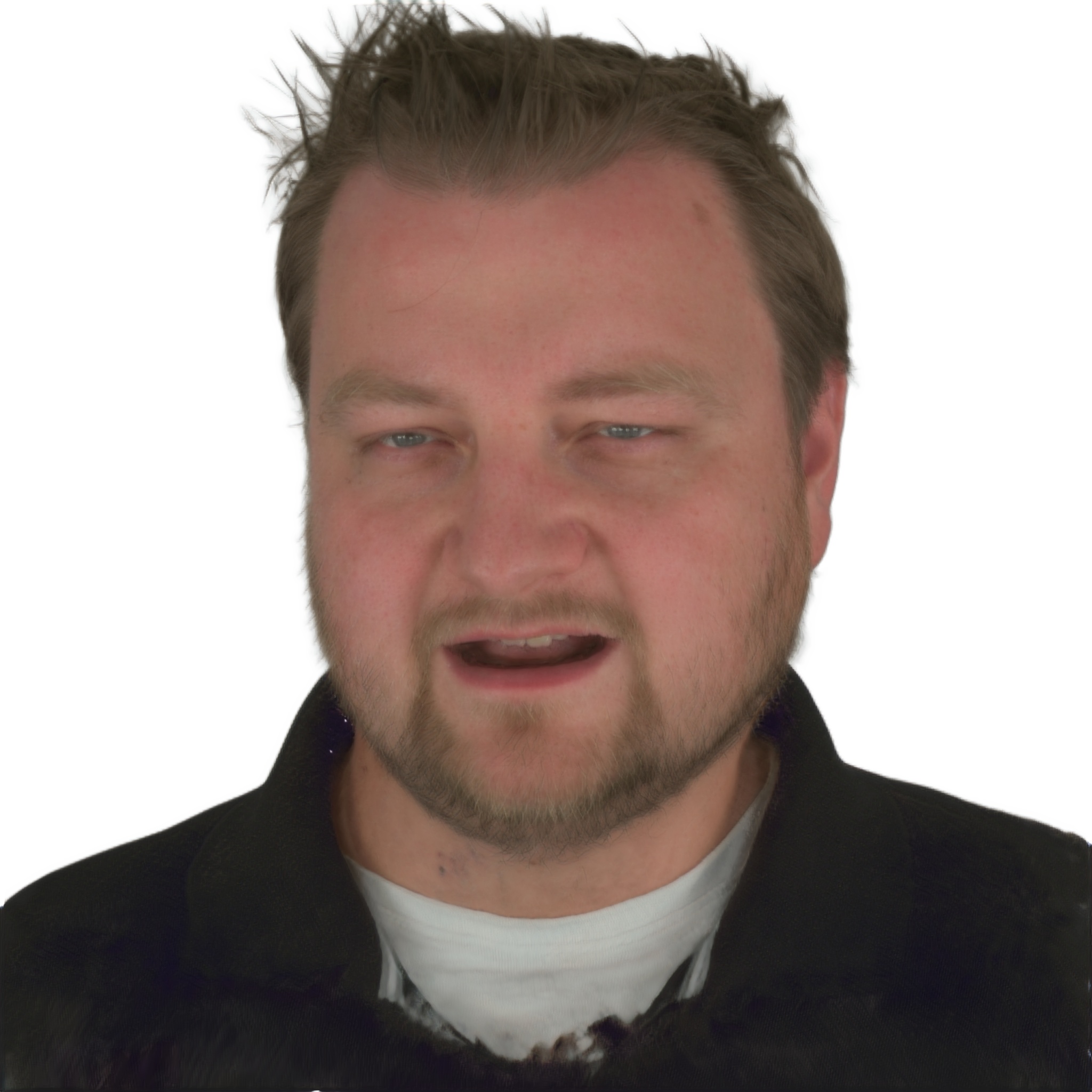}
         \caption{+Rand Noise}
         \label{fig:self_388_rand}
     \end{subfigure}
     \hfil
     \begin{subfigure}[b]{0.24\columnwidth}
         \centering
         \includegraphics[width=\textwidth]{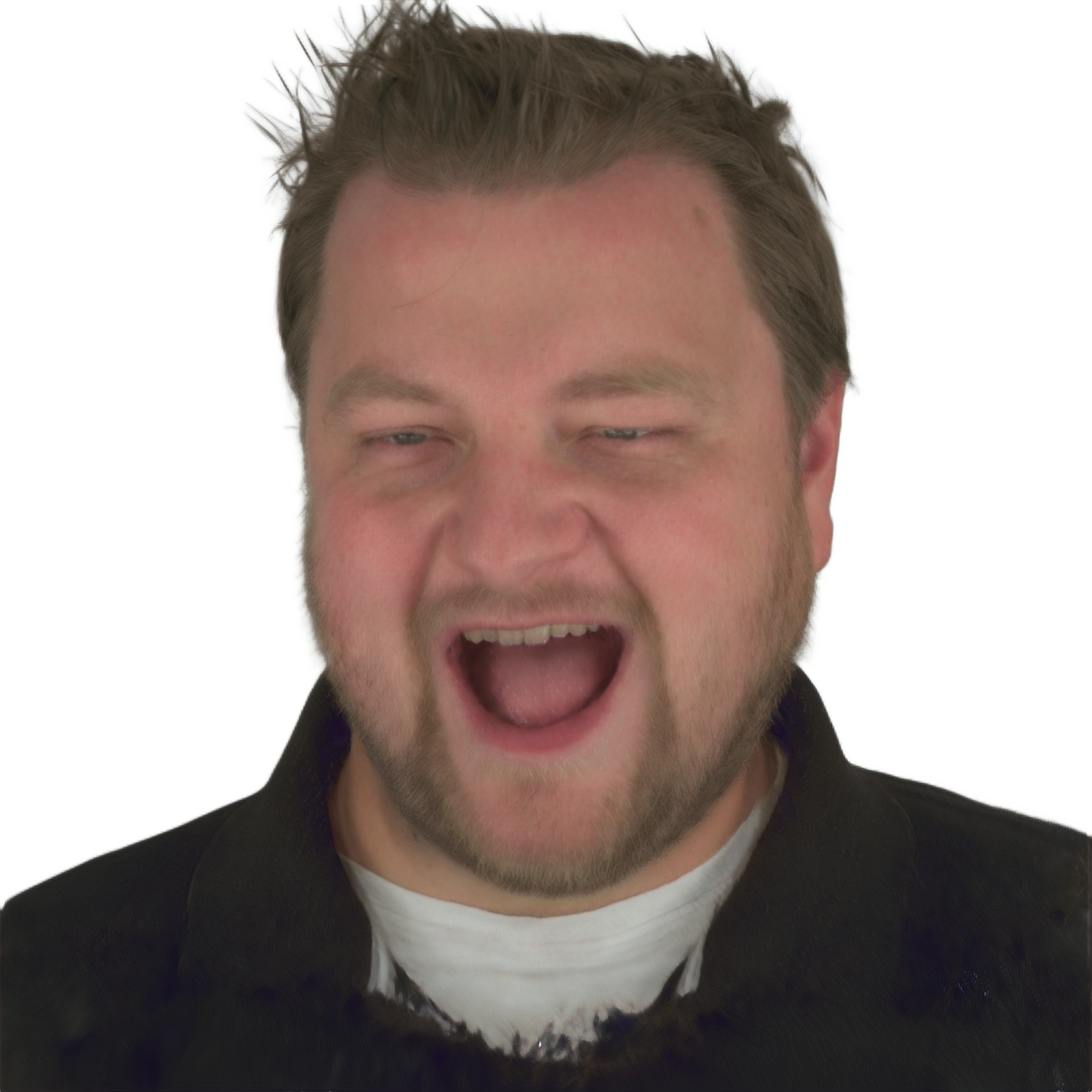}
         \caption{+\ours}
         \label{fig:self_388_ours}
     \end{subfigure}
     \vfil
     \begin{subfigure}[b]{0.24\columnwidth}
         \centering
         \includegraphics[width=\textwidth]{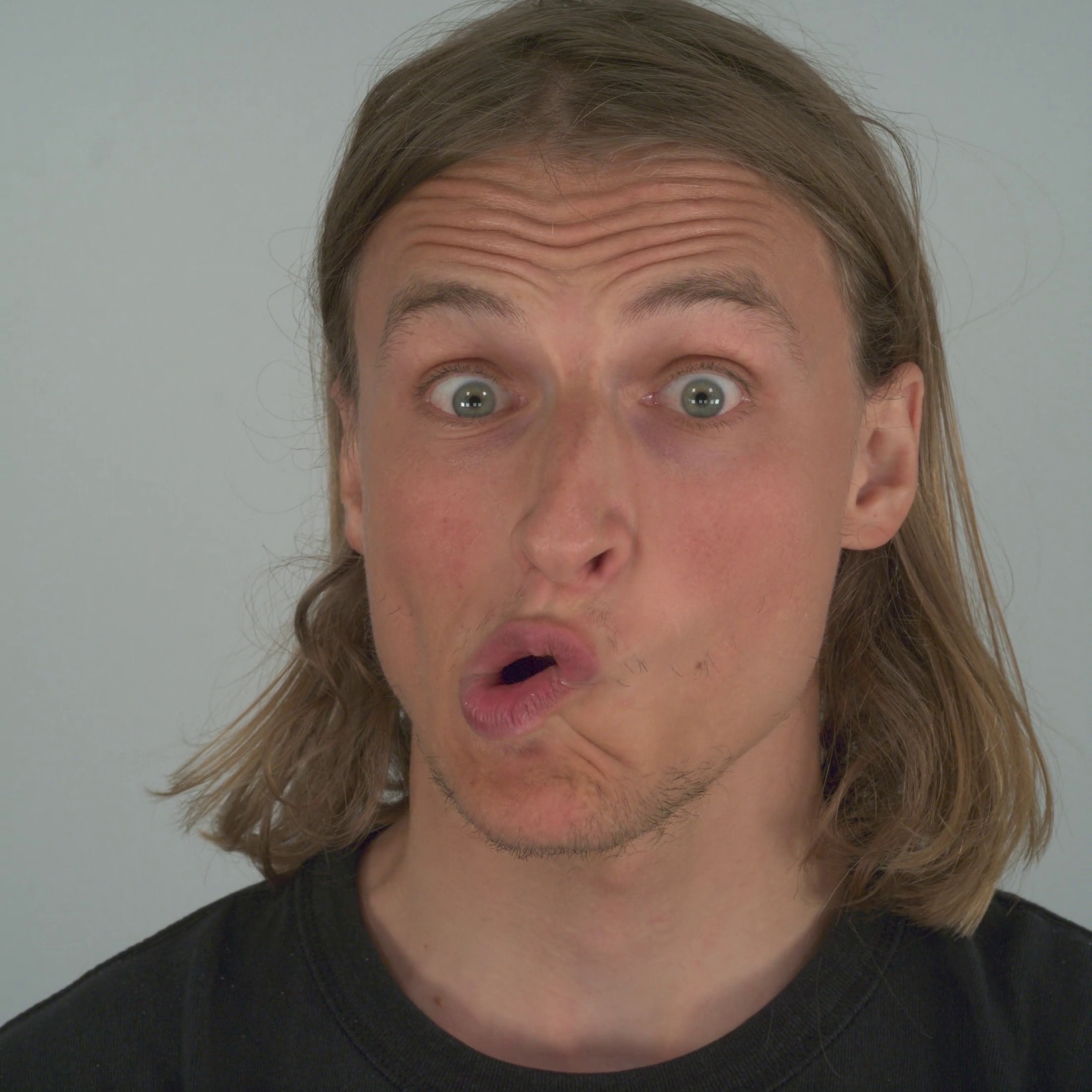}
         \caption{Driver}
         \label{fig:self_445_driver}
     \end{subfigure}
     \hfil
     \begin{subfigure}[b]{0.24\columnwidth}
         \centering
         \includegraphics[width=\textwidth]{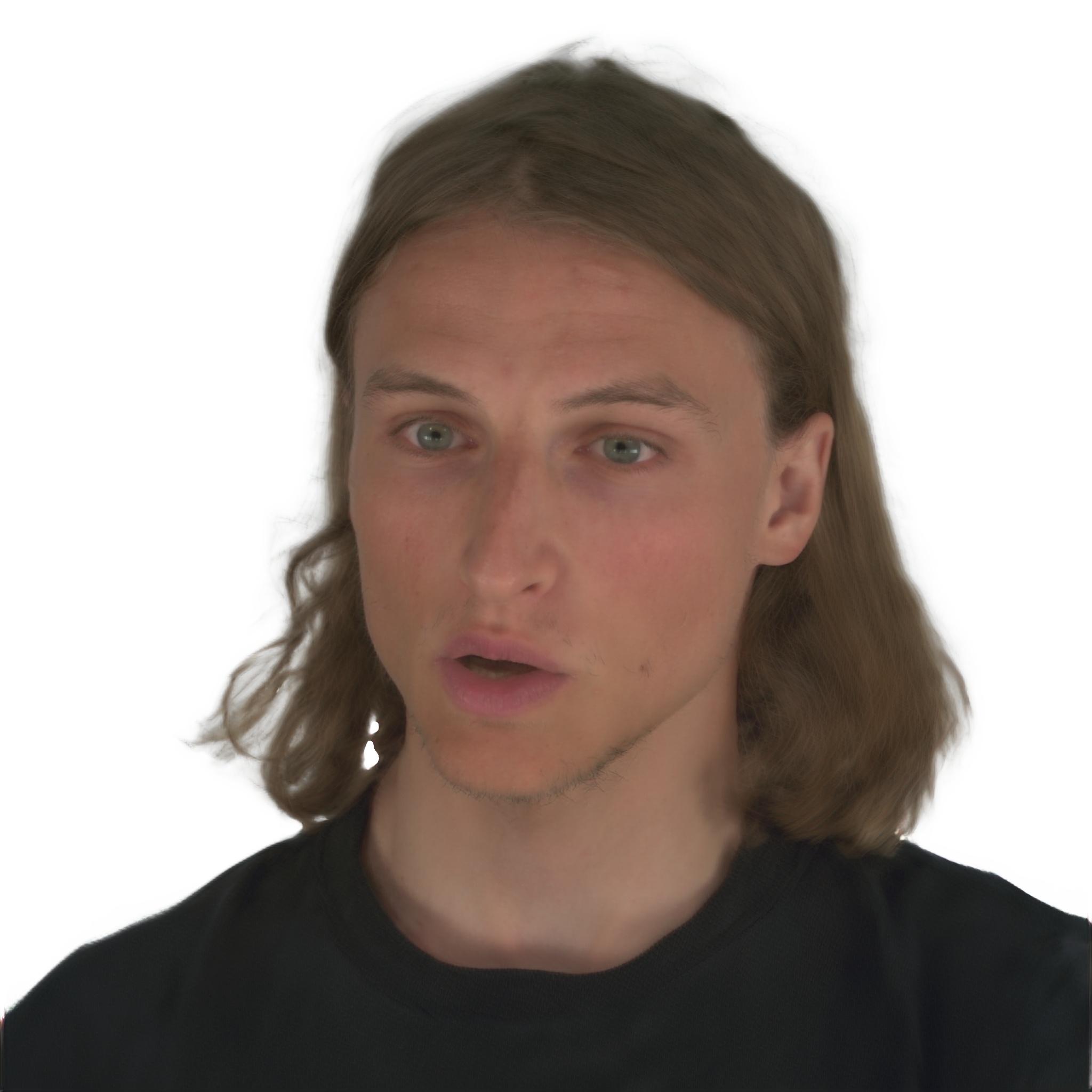}
         \caption{Original}
         \label{fig:self_445_baseline}
     \end{subfigure}
     \hfil
     \begin{subfigure}[b]{0.24\columnwidth}
         \centering
         \includegraphics[width=\textwidth]{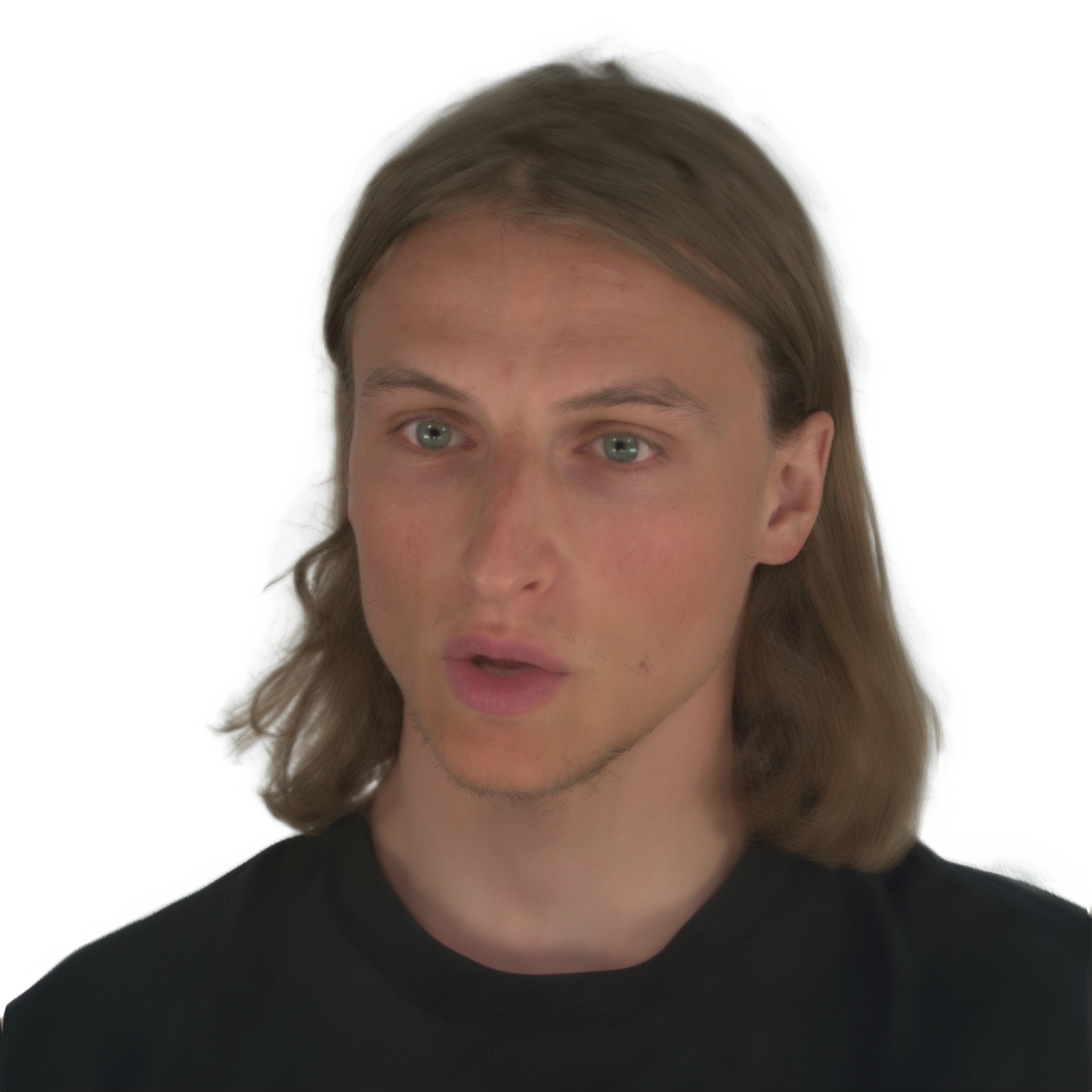}
         \caption{+Rand Noise}
         \label{fig:self_445_rand}
     \end{subfigure}
     \hfil
     \begin{subfigure}[b]{0.24\columnwidth}
         \centering
         \includegraphics[width=\textwidth]{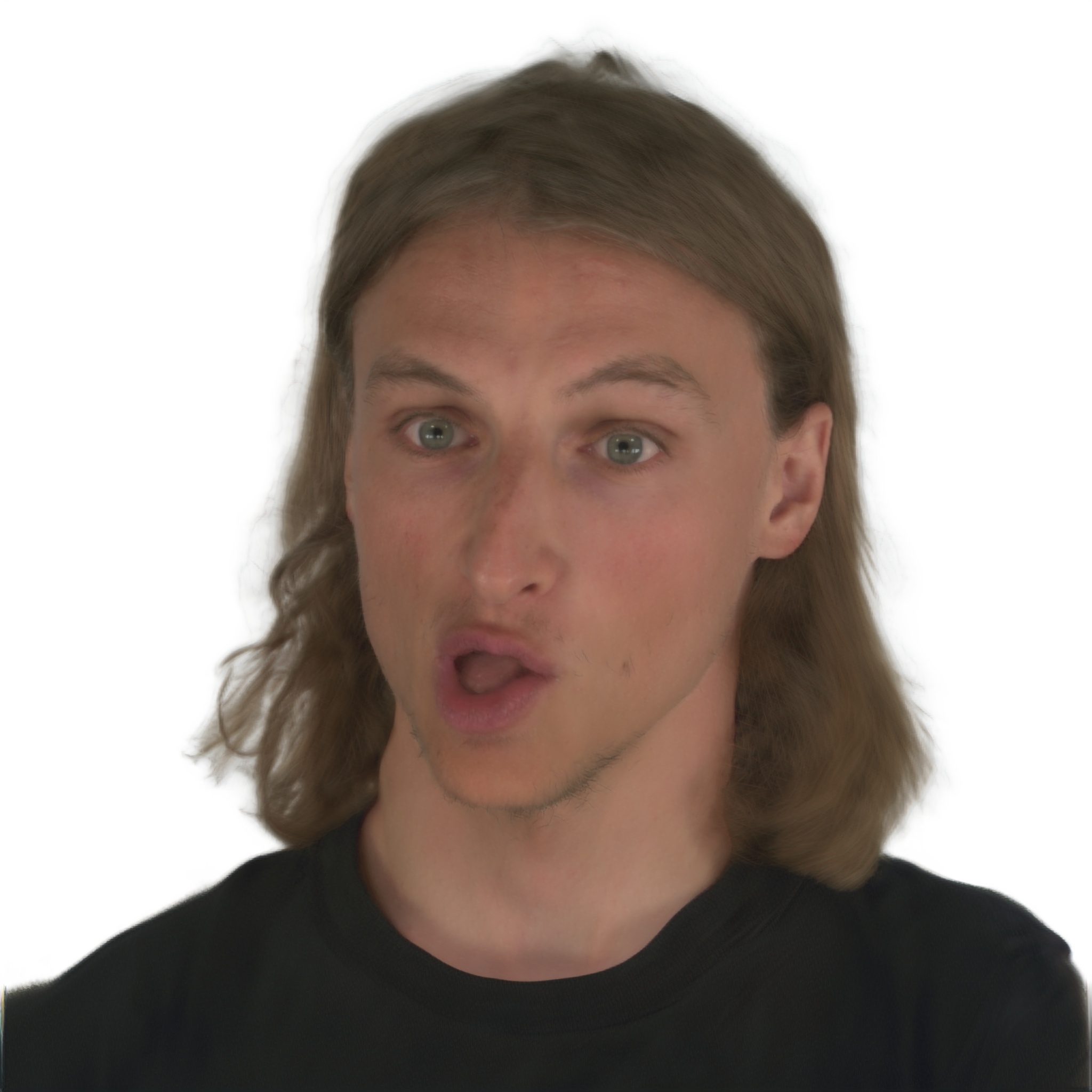}
         \caption{+\ours}
         \label{fig:self_445_ours}
     \end{subfigure}
\end{minipage}
}
     
  \caption[Images]{  
    Self-driving qualitative comparison. Compared to the baselines, \ours produces unseen expressions that more closely resemble those of the subject.} 
\label{fig:self_driving}
\end{figure*}

\textbf{Qualitative comparison: }
\Cref{fig:example_d445_a443} shows cross-driving results for three different avatars.  
The leftmost column contains the driver’s input frame, followed by the vanilla baseline~\cite{ultra_high_fid}, the Random Noise baseline, and our method.  
Across all identities and driving conditions, \ours more faithfully reproduces the driver’s facial expressions while better preserving the avatar’s identity and overall appearance.  
Notably, even in challenging cases such as the last row, where all methods struggle to match the driver’s fine-grained expression, \ours captures the driver’s underlying emotional state more convincingly - a trend further reflected in our quantitative emotion-similarity scores.

We also present self-driving results in \Cref{fig:self_driving}. 
Although the avatar never observes these expressions during training (as the ``FREE'' sequence is excluded), \ours is uniquely able to reproduce these unseen expressions more accurately than the baselines, demonstrating that expanding the expression space during training improves expressiveness even in the self-driving setting.

\textbf{Metrics: }
Following prior works~\cite{GAG,he2025lam}, we evaluate how accurately the rendered avatar reproduces the driver’s expression for both self- and cross-driving settings, determined by whether the driver and avatar share the same identity. We report the \emph{Average Expression Distance (AED)} and \emph{Average Pose Distance (APD)}, which measure the difference between the driver’s expression or head pose and the avatar’s reenacted output. Since cross-driving involves different identities, AED and APD are computed in the latent space of a pre-trained off-the-shelf 3DMM regressor (DECA~\cite{DECA}), which predicts FLAME expression and pose coefficients for both the driver and the generated frame; distances are then computed between the two coefficient vectors. Following Zielonka \etal \cite{zielonka2025gaussian}, we additionally measure \emph{emotion similarity} by extracting emotion features using a pre-trained EmoNet model~\cite{emonet}, which was trained for emotion classification of individuals. We compute the feature similarity between driver and reenacted frames. This metric captures high-level emotional consistency that is identity-agnostic and thus particularly meaningful in the cross-driving setting. For the self-driving scenario, where ground-truth frames are available, we also evaluate standard image-quality metrics, including \emph{PSNR} and \emph{SSIM}, following widely used protocols in recent avatar literature~\cite{chu2024generalizable,ye2024real3d,he2025lam}.

\textbf{Quantitative evaluation:}
As shown in \Cref{tab:results}, \ours outperforms both baselines across most metrics. In particular, \ours achieves substantial gains in AED and EmoNet similarity in both self- and cross-driving. Notably, although \ours augments cross-driving (other identities), it significantly improves self-driving performance. This is expected in our setting because the test split uses the ``FREE'' sequence, whose expressions are largely unseen during training. It further suggests that the bottleneck is not cross-identity per se, but expression coverage in the training distribution. As shown in \Cref{sec:motivation}, retrieval augmentation increases expression diversity and brings the effective training expression distribution closer to test drivers distribution, which improves generalization to the unseen ``FREE'' expressions.

In contrast, the \textit{Random Noise} baseline does not meaningfully improve expression reproduction. We also observe that \ours APD improves slightly in self-driving but worsens in cross-driving. We attribute this to pose/expression entanglement in the expression feature space: nearest neighbors retrieved using \emph{only} expression features tend to be similar not only in expression but also in head pose (\cref{sec:motivation}). During training-time augmentation, swapping in such features can inject pose-correlated cues that partially compete with the explicit pose conditioning, leading to small pose inconsistencies at test time and a higher APD. Conversely, \textit{Random Noise} yields the lowest APD, likely because perturbing expression features reduces their usable pose-correlated signal, encouraging the model to rely more strongly on the provided pose conditioning for head pose control.

\textbf{Implementation details:}
For each subject, we train the avatar on all available camera views at 25\,FPS for 500 epochs. The ``FREE'' sequence is excluded from training and used only for self-driving evaluation. We use Adam with a learning rate of $1\mathrm{e}{-4}$ and train on an NVIDIA A100 (40\,GB VRAM). To construct the expression bank, we sample approximately $200$ random frames per identity and extract their expression coefficients by fitting a BFM as in \cite{ultra_high_fid}. During evaluation, expression and pose distances (AED, APD) are computed using expression and pose coefficients from a pre-trained DECA regressor~\cite{DECA}, as standard metrics that differ from the training BFM space. Emotion similarity is measured using EmoNet~\cite{emonet} features. For cross-driving, each avatar is driven by the ``FREE'' sequence of all other benchmark subjects, supplemented with $10$ additional external drivers.

\textbf{Ablation study:}
We ablate the size and diversity of the expression bank used for nearest-neighbor retrieval in \ours (\Cref{sec:expression_bank}). Nearest-neighbor augmentation is most beneficial when the bank contains examples that are \emph{close enough} to provide meaningful supervision, yet \emph{different enough} to expand coverage. If the bank misses a relevant mode of expression (\eg raised eyebrows), retrieval can return suboptimal neighbors, weakening the augmentation signal. To quantify the effect of reduced bank diversity, we construct two restricted banks by retaining approximately half of the identities. As shown in \Cref{tab:ablations}, reducing the bank leads to only small changes in self-driving image-quality metrics, and a modest but consistent degradation in cross-driving metrics. This suggests that \ours is fairly robust once the bank is sufficiently diverse, while additional scale primarily improves the likelihood of finding high-quality neighbors for more challenging cross-identity motion-consistent with our user study in \Cref{sec:motivation}, where full-bank retrieval is only moderately preferred over retrieval from a much smaller bank. The gap between the two half-bank subsets indicates that expression \emph{coverage} matters more than the nominal size: subsets that better span the expression manifold yield retrievals that are closer to the full-bank performance. Overall, larger and more diverse banks remain preferable, but the gains exhibit diminishing returns once common expression variations are well covered. We further ablate over top-k candidates by sampling uniformly from the top-$5$ nearest neighbors (RAF top~5) replacing the default top-$1$. As shown in \Cref{tab:ablations}, top-$5$ sampling improves emotion similarity, but slightly degrades fine-grained motion accuracy and noticeably increases APD, revealing a trade-off between broader semantic/emotion alignment and precise expression/pose transfer.
\begin{table*}[h]
\centering
\small
\caption{Ablation study on a test set from the NeRSemble dataset~\cite{kirschstein2023nersemble}. The expression bank for nearest-neighbor retrieval was limited to approximately half of the identities.}
\label{tab:ablations}
\resizebox{\columnwidth}{!}{%
\begin{tabular}{lcccccccc}
\toprule
\multirow{2}{*}{Method} & 
\multicolumn{5}{c}{Self-Driving} & 
\multicolumn{3}{c}{Cross-Driving} \\ 
\cmidrule(lr){2-6} \cmidrule(lr){7-9}
& $\uparrow$ PSNR & $\uparrow$ SSIM & $\downarrow$ AED & $\downarrow$ APD & $\uparrow$ Emotion Sim. 
& $\downarrow$ AED & $\downarrow$ APD & $\uparrow$ Emotion Sim. \\ 
\midrule \rowcolor{lightgray}
Half bank \#1 & 18.070 & 0.7926 & 0.23513 & 0.13226 & 0.85169 & 0.27079 & 0.12495 & 0.80305 \\
Half bank \#2 & {\bf 18.137} & {\bf 0.7941} & 0.24693 & 0.13225 &  0.85616 & 0.26828 & {\bf 0.12295} & 0.80733 \\
RAF (top 5) & - & - & 0.23927 & 0.17058 & {\bf 0.88228} & 0.26687 & 0.15089 & {\bf 0.83073} \\
RAF & 18.002 & 0.7917 & {\bf 0.23094} & {\bf 0.12932} & 0.85595 & {\bf 0.26164} & 0.12687 & { 0.80801} \\
\bottomrule
\end{tabular}
}
  
\end{table*}
\vspace{-0.4cm}
\section{Discussion and Limitations}
Our \ours is designed for avatar methods that learn an expression-conditioned deformation field. When motion is inherited from a fixed 3DMM/LBS scaffold, retrieval-based expression substitution cannot substantially expand the avatar’s expressive range without changing the underlying template space. Our contribution should therefore be understood as improving the training regime of learned-deformation avatars rather than proposing a universal augmentation technique. \ours assumes that nearest neighbors in the chosen expression embedding correspond to perceptually similar facial motion. Imperfect tracking, identity-dependent biases, or mismatches between the feature space used for retrieval and the one used for evaluation can weaken the semantic meaning of ``nearest''. In addition, as our analysis suggests, expression embeddings may implicitly encode head-pose cues, and retrieval performed purely in expression space can inject pose-correlated signals that compete with explicit pose conditioning, slightly degrading pose accuracy.

\vspace{-0.3cm}
\section{Summary}
We present \ours, a retrieval-augmented training strategy that strengthens 3DGS head avatars that learn expression-driven deformation with MLPs. By substituting part of the subject’s expression features with nearest-neighbor expressions retrieved from a multi-identity bank, while still reconstructing the original target frame, \ours expands the effective expression supervision available to the deformation field. This simple plug-in improves robustness to unseen expressions and enhances cross-identity expression control without architectural changes, paired multi-identity supervision, or extra labels. Our results on the NeRSemble dataset show consistent gains in expression fidelity in both cross- and self-driving, supporting the view that expression coverage is a key bottleneck for ultra-high-fidelity learned-deformation avatars.

\ours demonstrates that even subject-specific Gaussian avatars can benefit from large-scale, cross-identity expression priors integrated purely at training time. This opens a promising avenue for future research: retrieval-augmented expression priors, scalable cross-identity supervision for 3DMM-free avatars, and richer multi-modal conditioning signals. More broadly, \ours suggests that coupling subject-specific reconstruction with broad, identity-agnostic expression control is a powerful direction for building robust, expressive, and widely deployable 3D head avatars.

\paragraph{\bf Acknowledgments:} 
This work was partly supported by the Israel Science Foundation (grants 2203/24).
%
%
\bibliographystyle{splncs04}
\bibliography{main}

\appendix
\newpage
\section*{{\huge Supplementary Material}}
This supplementary material provides additional details and supporting analysis for our paper. In \Cref{supp:metrics}, we elaborate on the distribution-level evaluation metrics used to quantify expression coverage and similarity. In \Cref{supp:user_study}, we describe the setup and protocol of our Amazon Mechanical Turk user study, including worker qualifications and annotation tasks. Finally, in \Cref{supp:impl}, we provide further implementation details, including the random-noise baseline and the benchmark identities used in our experiments.

\section{Evaluation Metrics Details}
\label{supp:metrics}

In \Cref{tab:bank_analysis} of the main paper, we report three distribution-level metrics to measure how well each subject's training expression features cover the unseen driver test distribution, both before (\textit{Vanilla}) and after applying retrieval augmentation (\textit{+RAF}).
We describe each metric below.

\paragraph{Maximum Mean Discrepancy (MMD).}
MMD quantifies the statistical distance between two sets of samples using the kernel trick. Concretely, we compare the subject's training expression features, either the original set (\textit{Vanilla}) or the augmented set (\textit{+RAF}), against the full set of unseen driver expression features.
We employ an RBF (Gaussian) kernel, so MMD measures how closely the two distributions match in a reproducing kernel Hilbert space. Lower MMD values indicate greater distributional similarity between the subject's training expressions and the target driver distribution.

\paragraph{Kullback-Leibler (KL) Divergence.}
KL divergence measures how much information is lost when one probability distribution is used to approximate another.
We estimate $\mathrm{KL}(P_{\text{subject}} \| P_{\text{target}})$ via kernel density estimation (KDE) in a PCA-reduced expression-feature space.
$P_{\text{subject}}$ is the density of the subject's training expression features (Vanilla or +RAF) and $P_{\text{target}}$ is the density of the unseen driver expression features.
Lower KL values indicate that the subject's training distribution is a better fit to the target distribution.

\paragraph{Bank-to-Train Nearest-Neighbor Distance (B2T).}
B2T measures \emph{coverage}: for every expression vector in the unseen driver set (target), we find its closest neighbor in the subject's training set (Vanilla or +RAF) and compute the Euclidean distance.
The reported value is the mean of these per-sample minimum distances.
Lower B2T values indicate that the subject's training set covers the target expression manifold more densely, \ie unseen driver expressions are never far from a training-set neighbor, reflecting greater expression diversity and representativeness.


\section{User Study Details}
\label{supp:user_study}

We conducted the user study on Amazon Mechanical Turk (AMT).
To ensure high annotation quality, we applied the following worker filters:
\begin{itemize}
    \item \textbf{Geographical restriction:} Workers from English-speaking countries only (United States, United Kingdom, Australia, Canada, and India).
    \item \textbf{Approval rate:} HIT approval rate of 98\% or above.
    \item \textbf{Masters Qualification:} Only workers holding the AMT Masters Qualification were eligible.
    According to Amazon Mechanical Turk, Masters are workers who \textit{``have demonstrated excellence across a wide range of tasks''} and are continuously monitored to retain the qualification.
\end{itemize}
In total, \textbf{36 unique workers} participated in the study, collectively annotating 1{,}000 query images.

\paragraph{Task Description.}
Each Human Intelligence Task (HIT) presented workers with \textbf{two pairs of face images} displayed side by side: a \emph{Left Pair} (Reference + Candidate~A) and a \emph{Right Pair} (Reference + Candidate~B), where the same Reference image appeared in both pairs.
Workers were explicitly instructed to ignore the identity of the depicted individuals.
They answered two forced-choice questions:
\begin{enumerate}
    \item \textbf{Q1 Head pose:} ``Which pair matches the \textbf{head pose} of the reference image better?'' \\
    \textit{Note: Ignore facial expression here, look only at head orientation.}
    \item \textbf{Q2 Facial expression:} ``Ignoring identity, which pair matches the \textbf{facial expression} of the reference image better?'' \\
    \textit{Note: Ignore head orientation here, look only at the expression \\(mouth/eyes/eyebrows).}
\end{enumerate}
For each question, response options were \emph{Left Pair}, \emph{Right Pair}, or \emph{Tie / Can't tell}.
We evaluated two comparisons: (1)~the nearest neighbor (NN) retrieved from the full expression bank \vs a random bank image, and (2)~the NN from the full bank \vs the NN retrieved from a much smaller bank.

\begin{figure}[h]
    \centering
    \includegraphics[width=\linewidth]{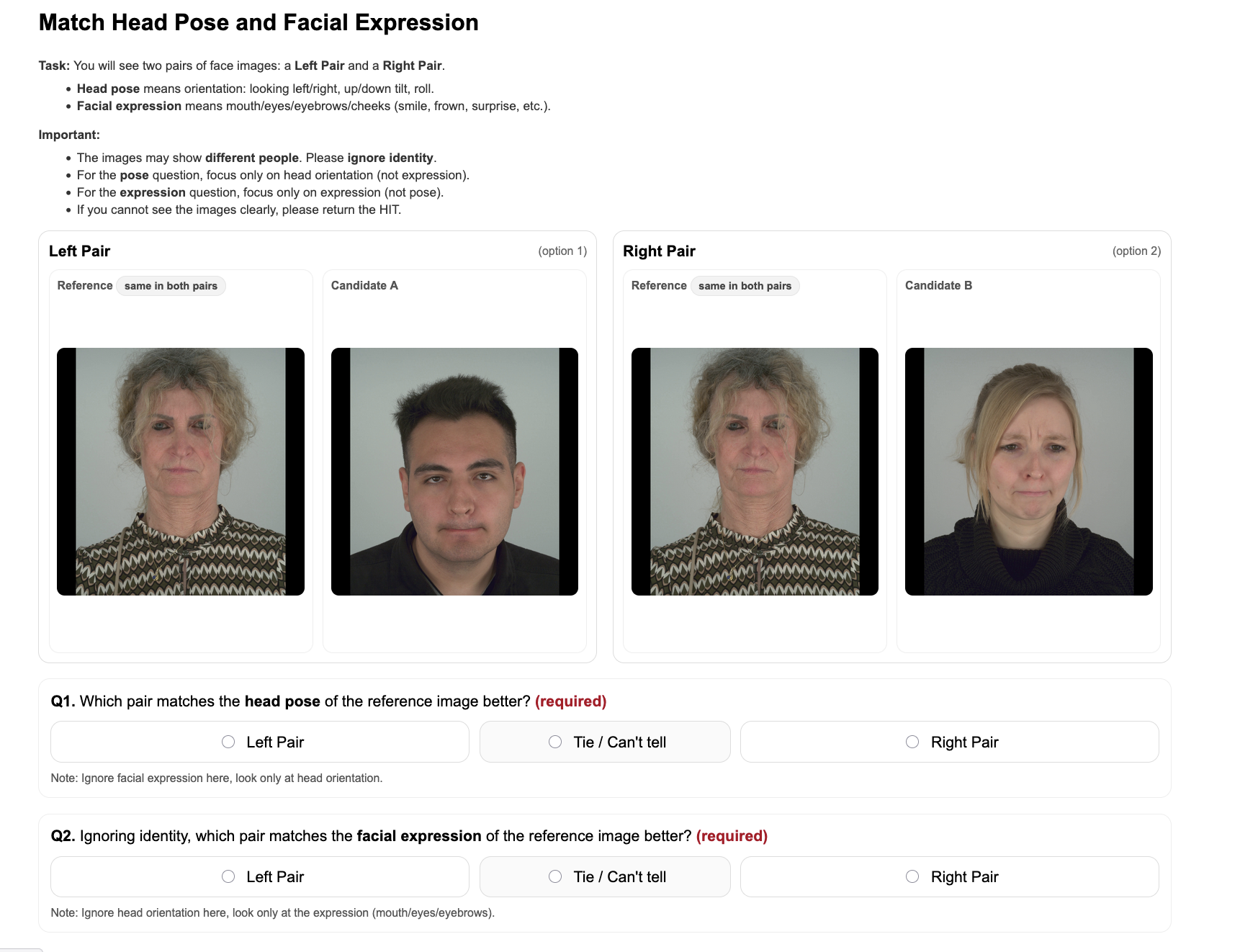}
    \caption{Screenshot of the Amazon Mechanical Turk (AMT) user study interface.
    Each HIT presents two pairs of face images (Left Pair and Right Pair), each consisting of a Reference image and a Candidate.
    Workers answer two forced-choice questions---one on head pose similarity and one on facial expression similarity---while being instructed to ignore identity.}
    \label{fig:amt_screenshot}
\end{figure}


\section{Additional Implementation Details}
\label{supp:impl}

\paragraph{Random Noise Baseline.}
In the \textit{Random Noise} baseline, expression features are augmented at training time by adding Gaussian noise.
The noise standard deviation is set to \textbf{0.08}, which is on the order of the average $\ell_2$ norm of the expression feature vectors, making the perturbation non-trivial.
We also experimented with a larger noise magnitude of 0.5, which showed similar qualitative trends; notably, the larger noise yielded even lower APD scores.
As discussed in the main paper, this is attributed to the fact that strong noise effectively destroys any pose-correlated signal that may leak through the expression feature channel, forcing the network to rely exclusively on the dedicated pose feature for head-pose control.
The reduction in APD is therefore a byproduct of reduced expression-pose entanglement rather than improved expression fidelity.

\paragraph{Benchmark Identities.}
Following the standard NeRSemble benchmark~\cite{kirschstein2023nersemble}, we evaluate on five subjects of the following IDs: \textbf{388}, \textbf{422}, \textbf{443}, \textbf{445}, and \textbf{475}.
\end{document}